\pdfoutput=1
\documentclass[wcp]{jmlr}

\usepackage{amsmath,amsfonts,bm}

\def\eqref#1{equation~\ref{#1}}

\def\1{\bm{1}}

\DeclareMathAlphabet{\mathsfit}{\encodingdefault}{\sfdefault}{m}{sl}
\SetMathAlphabet{\mathsfit}{bold}{\encodingdefault}{\sfdefault}{bx}{n}

\usepackage{packages}

\newcommand{\ours}{{\texttt{OPTIMA}}\xspace}
\newcommand{\our}{{\texttt{OPTIMA}}}
\definecolor{shade}{RGB}{220, 230, 255}
\setlist[itemize]{
  noitemsep,
  topsep=0pt,
  leftmargin=*
}
\newcommand{\columndistance}{2.0cm}

\newif\ifshort
\ifdefined\shortversion\shorttrue\else\shortfalse\fi
\newcommand{\fullversionurl}{https://arxiv.org/pdf/2512.13886}
\ifshort
  \usepackage{xr}
  \usepackage{refcount}
  \makeatletter
  \newcommand{\appref}[1]{%
    \edef\appref@url{\fullversionurl\string##page=\getpagerefnumber{full-#1}}%
    \href{\appref@url}{Appendix~\getrefnumber{full-#1} of the full version}}
  \makeatother
\else
  \newcommand{\appref}[1]{\autoref{#1}}
\fi

\definecolor{KeywordColor}{RGB}{0, 102, 153}
\definecolor{VarColor}{RGB}{128, 0, 128}
\definecolor{CommentColor}{RGB}{0, 128, 0}

\renewcommand{\algorithmiccomment}[1]{\hfill {\color{CommentColor}\scriptsize$\triangleright$ #1}}

\algrenewcommand\alglinenumber[1]{\scriptsize #1}

\newcommand{\doublemidrule}{\midrule[\heavyrulewidth]
                            \midrule[\heavyrulewidth]}

\ifshort\else\fi

\makeatletter
\let\Ginclude@graphics\@org@Ginclude@graphics
\makeatother

\jmlrvolume{353,}
\jmlryear{2026}
\jmlrworkshop{ACML 2026}

\title[\ours: Optimal One-Shot LLM Pruning via QP Reconstruction]{%
  \ours: Optimal One-Shot Pruning for LLMs via Quadratic Programming Reconstruction}

\author{\Name{Mohammad Mozaffari\nametag{$^{1}$\thanks{Equal contribution.}}} \Email{mmozaffari@cs.toronto.edu}\\
  \Name{Samuel Kushnir\nametag{$^{2}$\footnotemark[1]}} \Email{samuelkushnir@google.com}\\
  \Name{Maryam Mehri Dehnavi\nametag{$^{1}$}} \Email{mmehride@cs.toronto.edu}\\
  \Name{Amir Yazdanbakhsh\nametag{$^{2}$\thanks{Contributed to this paper in an advisory capacity.}}} \Email{ayazdan@google.com}\\
  \addr $^{1}$University of Toronto \qquad $^{2}$Google DeepMind}

\editors{Andy Song, Bo Han and Sarah Erfani}

\begin{document}

\makeatletter
\let \@jmlrpages \@empty
\makeatother

\maketitle

\begin{abstract}
Post-training pruning cuts the inference cost of large language models, yet it faces a trade-off: \textit{simple heuristics that zero weights are fast but degrade accuracy, while principled joint optimization methods recover accuracy but are computationally infeasible at modern scale}.
One-shot methods such as SparseGPT sit in between and apply efficient but approximate weight updates. To close this gap, we introduce \ours, a practical one-shot post-training pruning method that balances accuracy and scalability.
\ours casts layer-wise weight reconstruction after mask selection as independent, column-wise quadratic programs (QPs) that share one Hessian per weight matrix.
Solving these QPs yields, at solver convergence, the per-column minimizer of the layer-wise reconstruction objective under the shared Hessian, and the shared Hessian makes the QPs easy to batch on accelerators.
We implement an accelerator-friendly QP solver that accumulates one Hessian per weight matrix and solves many small QPs in parallel with a convergence certificate per column, enabling one-shot post-training pruning at scale on a single accelerator without fine-tuning.
\ours integrates with existing mask selectors and improves average zero-shot accuracy in 40 of 49 (model, mask selector, sparsity) settings across the LLaMA and Gemma families, with mean gains of $+0.61$, $+0.56$, and $+1.01$ points at 50\% unstructured, 2:4, and 60\% unstructured sparsity, and up to $+3.97$ points at 60\% sparsity, while matching ADMM-based exact reconstruction on its own masks.
On an NVIDIA H100, \ours prunes an 8B-parameter transformer end-to-end in 40 hours and 60 GB of peak GPU memory. This shows that principled column-wise QP reconstruction is a practical alternative to heuristic weight updates in one-shot post-training pruning.\footnote{Code and data: \url{https://github.com/paramathic/optima}}
\end{abstract}

\begin{keywords}
Large language models; post-training pruning; model compression; quadratic programming; sparsity.
\end{keywords}

\section{Introduction}
\label{sec:intro}
Large language models (LLMs) deliver strong capabilities across a wide array of natural language tasks~\citep{gemini, gemini2.5, llama2, deepseekr1}.
However, growing parameter counts make deployment and inference expensive.
Post-training one-shot pruning~\citep{sparsity_survery}, which removes parameters from a pretrained model with only a small calibration dataset, promises to reduce these costs, yet it faces a fundamental trade-off: fast, heuristic schemes that simply zero weights (e.g., Wanda~\citep{wanda}) are cheap but often incur noticeable accuracy losses, while the full inverse-Hessian update of Optimal Brain Surgeon~\citep{obs} recovers accuracy but is infeasible at LLM scale, and its scalable approximations \citep{obc, llmsurgeon} trade optimality for cost.
One-shot approximations such as SparseGPT~\citep{sparsegpt} and related heuristics~\citep{thanos} try to navigate this middle ground, but they sacrifice reconstruction optimality and therefore leave headroom in accuracy.

\begin{figure}
    \centering
    \includegraphics[width=0.5\linewidth]{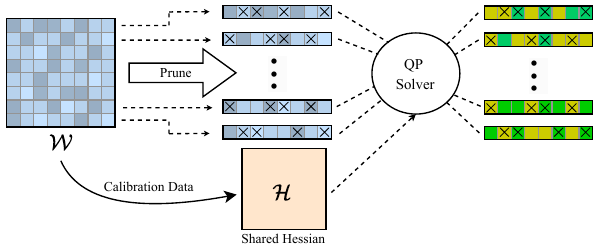}
    \caption{Overview of \ours. One Hessian $\mathbf{H}$ per weight matrix is estimated from the calibration data; after the mask selector fixes pruned entries, each column's remaining weights are updated by a small QP with $\mathbf{H}$, and the identically structured QPs are solved in parallel on an accelerator.}
    \label{fig:optima}

    \vspace{-20pt}
\end{figure}

In this paper we introduce \ours, a practical one-shot post-training pruning framework that closes much of this gap by combining principled optimality with batched, certified solves on accelerators.
The core idea (\autoref{fig:optima}) is a reformulation of the layer-wise reconstruction step that follows mask selection.
After fixing a binary mask for a weight matrix, the reconstruction (least-squares) objective decomposes across columns and each column's update can be written as a small quadratic program (QP).
Crucially, every column of the same weight matrix shares the same Hessian matrix $\mathbf{H} = \mathbf{X}^T \mathbf{X}$, while the linear constraints differ only according to which entries the mask removes.
This shared-Hessian, column-wise QP structure yields two immediate benefits: (1) per-column global optimality for the reconstruction objective (given the estimated Hessian), and (2) uniform problem structure that enables batching and parallelism on off-the-shelf ML accelerators.

Realizing this formulation in practice requires careful numerical and systems engineering. We adopt a first-order primal-dual QP solver (restarted accelerated primal-dual hybrid gradient, rAPDHG~\citep{pdqp}) that is well-suited to our constrained problems and whose critical operations reduce to matrix-vector products with the shared Hessian.
This keeps the inner loop to matrix-vector products on accelerators.
We further avoid explicit dense equality matrices by enforcing fixed entries via tight bounds, accumulate per-matrix Hessians incrementally from calibration sequences to save memory, and solve columns in batches so thousands of small QPs are processed in parallel.
These implementation choices make \ours not only theoretically principled but also practical to run on a single accelerator.

Solving the post-mask reconstruction exactly is not new in itself: \citet{boza_admm} solve it with the alternating direction method of multipliers (ADMM), and ALPS \citep{alps} refines its own ADMM-selected support with a vectorized GPU conjugate-gradient solve on the shared Hessian. Our contribution is a bound-constrained primal-dual formulation that terminates with a convergence certificate per column, and its evaluation as a drop-in update on masks from five selectors (Wanda, SparseGPT, Thanos, ProxSparse, and ADMM).

We evaluate \ours across multiple model families (LLaMA, Gemma, and Qwen) and sparsity regimes (unstructured and 2:4 semi-structured sparsity).
\ours is modular and plugs into existing mask selectors (e.g., Wanda, SparseGPT, Thanos), improving zero-shot performance in most settings.
Across six zero-shot downstream benchmarks in Language Model Evaluation Harness, we observe up to 3.97 points of average accuracy gain on LLaMA and Gemma (4.19 on Qwen 2.5) without any post-pruning fine-tuning. In summary, our contributions are:
\begin{itemize}
\item We cast the post-training reconstruction problem as batched column-wise QPs that share one Hessian per weight matrix, equivalent to the least-squares objective after mask selection and solved to a per-column optimum (\autoref{sec:methodology}).
\item We design and implement an accelerator-friendly QP solver pipeline that accumulates one Hessian per weight matrix, enforces mask constraints via bounds, batches thousands of column QPs, and leverages rAPDHG/MPAX on accelerators (\autoref{alg:layerwise_pruning}).
\item We show the modularity of \ours, which can be used as a drop-in weight-update step with common mask selection algorithms (Wanda, SparseGPT, Thanos), improving their average accuracy without fine-tuning in 40 of 49 (model, mask selector, sparsity) settings (\autoref{sec:results}).
\item We provide extensive empirical evidence and practical measurements. \ours yields average accuracy gains of up to 3.97 points on LLaMA and Gemma, largest at higher sparsity, matches the ADMM-based exact reconstruction of \citet{boza_admm} on its own masks, and prunes an 8B model on a single H100 in about 40 hours and a 70B model in about 350 hours.
\end{itemize}

\section{Preliminaries}
\label{sec:preliminaries}

Post-training pruning (PTP) compresses pre-trained models without retraining, using a small calibration dataset to produce a sparse model that preserves performance.
To make PTP tractable, the problem is decomposed into independent layer-wise subproblems. For layer $l$, the goal is to find a binary sparsity mask $\mathbf{M}_l$ and updated weights $\hat{\mathbf{W}}_l$ that minimize the output reconstruction error given original weights $\mathbf{W}_l$ and input activations $\mathbf{X}_l$. This task can be formulated as in \autoref{eq:joint_optimization}, where $\odot$ denotes the Hadamard product, and $\mathbf{M}_l$ is a binary tensor of the same shape as $\mathbf{W}_l$ with 0s for pruned weights and 1s for retained ones. \autoref{eq:joint_optimization} is solved sequentially across layers, with $\mathbf{X}_l$ as the pruned output from layer $l-1$. Finding the optimal $\mathbf{M}_l$ is NP-hard, motivating heuristics.

\begin{equation}
    \underset{\mathbf{M}_l, \hat{\mathbf{W}}_l}{\text{argmin}} \|\mathbf{X}_l \mathbf{W}_l - \mathbf{X}_l(\mathbf{M}_l \odot \hat{\mathbf{W}}_l)\|_F^2
    \label{eq:joint_optimization}
\end{equation}

A common heuristic decouples mask selection from weight updates. After selecting $\mathbf{M}_l$ (e.g., by magnitude), the problem simplifies to \autoref{eq:weight_only}, which is a convex least-squares problem, but solving it directly is computationally expensive for large LLM weights.

\begin{equation}
    \min_{\hat{\mathbf{W}}_l} \|\mathbf{X}_l \mathbf{W}_l - \mathbf{X}_l(\mathbf{M}_l \odot \hat{\mathbf{W}}_l)\|_F^2
    \label{eq:weight_only}
\end{equation}

Consequently, many methods employ strategies to circumvent the expensive weight update step. For example, \textbf{Wanda} \citep{wanda} avoids weight updates altogether, simply setting the selected weights to zero. However, other methods such as \textbf{SparseGPT} \citep{sparsegpt} and \textbf{Thanos} \citep{thanos} adopt a compromise, performing a more complex update but only on a small subset of the weights. These heuristics trade off optimality for computational feasibility.

\section{\our: Optimal weight updates via quadratic programming}
\label{sec:methodology}
\ours updates \textbf{all} remaining weights once the pruning mask $\mathbf{M}_l$ has been chosen, by reformulating the least-squares problem as a set of independent QPs that are solved in parallel on accelerators with iterative methods.

\subsection{Reformulation as a quadratic program with linear constraints}
\label{sec:qp}

As discussed in \autoref{sec:preliminaries}, our goal is to minimize the problem defined in \autoref{eq:weight_only}. The Frobenius norm objective function in \autoref{eq:weight_only} is separable by the columns of the weight matrix. We can therefore solve the optimization problem for each column independently.

Let $\mathbf{w}_j$ be the $j$-th column of the original weight matrix $\mathbf{W}_l$, and let $\hat{\mathbf{w}}_j$ be the corresponding column in the updated matrix $\hat{\mathbf{W}}_l$. The mask for this column is $\mathbf{m}_j$. The optimization for this single column is $\min_{\hat{\mathbf{w}}_j} \|\mathbf{X}_l \mathbf{w}_j - \mathbf{X}_l (\mathbf{m}_j \odot \hat{\mathbf{w}}_j)\|_2^2$. By defining the change in the weight column as $\Delta\mathbf{w}_j = (\mathbf{m}_j \odot \hat{\mathbf{w}}_j) - \mathbf{w}_j$, the objective can then be rewritten in terms of this change as in \autoref{eq:delta_w} in standard quadratic form.

\begin{equation}
\min_{\Delta\mathbf{w}_j} \|-\mathbf{X}_l \Delta\mathbf{w}_j\|_2^2 = \min_{\Delta\mathbf{w}_j} \Delta\mathbf{w}_j^T (\mathbf{X}_l^T \mathbf{X}_l) \Delta\mathbf{w}_j
\label{eq:delta_w}
\end{equation}

The constraints on $\Delta\mathbf{w}_j$ in \autoref{eq:delta_w} are determined by the mask $\mathbf{m}_j$. Let $\mathcal{S}_j$ be the set of indices where the mask is zero (i.e., weights to be pruned). For each index $i \in \mathcal{S}_j$, the corresponding entry in the updated weight vector, $(\hat{\mathbf{w}}_j)_i$, must be zero. This imposes a linear constraint on the change vector, as shown in \autoref{eq:constraints}.

\begin{equation}
(\mathbf{m}_j \odot \hat{\mathbf{w}}_j)_i = 0 \implies (\Delta\mathbf{w}_j)_i = -(\mathbf{w}_j)_i \quad \forall i \in \mathcal{S}_j
\label{eq:constraints}
\end{equation}

The entries of $\Delta\mathbf{w}_j$ for the unpruned weights (where $m_{ij}=1$) remain as free variables to be optimized.

For each column $j$ of the weight matrix, we have a QP of the form represented in \autoref{eq:constrained_objective}, where $\mathbf{H} = \mathbf{X}_l^T \mathbf{X}_l$ is the Hessian matrix, which is positive semi-definite and shared across all column-wise problems. The fact that the Hessian is shared among all columns, and only the constraints change, makes the QPs straightforward to batch on accelerators.

\begin{equation}
\begin{aligned}
& \underset{\Delta\mathbf{w}_j}{\text{minimize}}
& & \Delta\mathbf{w}_j^T \mathbf{H} \Delta\mathbf{w}_j \\
& \text{subject to}
& & (\Delta\mathbf{w}_j)_i = -(\mathbf{w}_j)_i, \; \forall i \in \mathcal{S}_j
\end{aligned}
\label{eq:constrained_objective}
\end{equation}

Optimality here is defined with respect to the estimated Hessian and the chosen mask; the minimizer is unique when $\mathbf{H}$ restricted to the free coordinates is positive definite, and $\mathbf{H}$ is not damped. It does not imply global optimality of the joint mask-and-weight problem in \autoref{eq:joint_optimization}, nor does it guarantee better end-to-end task accuracy than heuristic updates.

The column-wise problem also admits an \emph{unconstrained} reformulation: substituting the fixed pruned entries directly into the objective yields a smaller QP over only the unpruned indices. We adopt the constrained form in our implementation because the unconstrained problems have column-dependent dimensions, which complicates batching on accelerators. The full derivation is in \appref{app:unconstrained}.

\subsection{Solving the quadratic programs}

We solve the QPs with the rAPDHG method \citep{pdqp}, a first-order method effective here for three reasons: (1) its bottleneck, matrix-vector multiplications with $\mathbf{H}$ and its transpose, runs efficiently on accelerators; (2) it achieves provably optimal linear convergence; and (3) an open-source JAX implementation is available in MPAX \citep{mpax}, which runs on GPUs and TPUs. This enables parallel solving of thousands of column-wise QPs, leveraging the shared structure.

\subsection{Efficient implementation}

Solving \autoref{eq:constrained_objective} naively costs too much time and memory. Four techniques make it fit on one GPU even for large LLMs, and \autoref{alg:layerwise_pruning} gives the full procedure.

\begin{algorithm}[t]
\caption{\textbf{Layer-wise pruning with batched column-wise quadratic programming}}
\label{alg:layerwise_pruning}
\begin{scriptsize}
\begin{algorithmic}[1]
\Statex \textbf{\textcolor{KeywordColor}{Input:}} Pre-trained LLM $\textcolor{VarColor}{\mathcal{M}}$, calibration data $\textcolor{VarColor}{\mathbf{X}}$, pruning masks $\textcolor{VarColor}{\mathcal{M}_{\text{ask}}}$, QP solver $\textcolor{VarColor}{\texttt{QPSolve}}$, batch size $\textcolor{VarColor}{B}$.
\Statex \textbf{\textcolor{KeywordColor}{Output:}} Pruned and updated LLM $\textcolor{VarColor}{\hat{\mathcal{M}}}$.

\vspace{0.5em}
\For{\textbf{each layer} $\textcolor{VarColor}{L}$ in the LLM $\textcolor{VarColor}{\mathcal{M}}$ \textbf{(sequentially)}}
    \State Capture the inputs $\{\textcolor{VarColor}{x_{\mathbf{W},i}}\}_{i=1}^{b}$ of every weight matrix $\textcolor{VarColor}{\mathbf{W}}$ in $\textcolor{VarColor}{L}$ from a forward pass of $\textcolor{VarColor}{L}(\textcolor{VarColor}{\mathbf{X}})$, one sequence at a time.
    \vspace{0.3em}
    \For{\textbf{each weight matrix} $\textcolor{VarColor}{\mathbf{W}}$ in layer $\textcolor{VarColor}{L}$}
        \State Initialize Hessian estimate $\textcolor{VarColor}{\mathbf{H}_{\mathbf{W}}} \gets 0$. \algorithmiccomment{One covariance matrix per weight matrix}
        \For{\textbf{each calibration sequence} $i = 1, \dots, b$}
            \State $\textcolor{VarColor}{\mathbf{H}_{\mathbf{W}}} \gets \textcolor{VarColor}{\mathbf{H}_{\mathbf{W}}} + \frac{1}{bs}\textcolor{VarColor}{x_{\mathbf{W},i}^T} \textcolor{VarColor}{x_{\mathbf{W},i}}$ \algorithmiccomment{Token-mean covariance of the inputs to $\mathbf{W}$}
        \EndFor
        \State Retrieve corresponding mask $\textcolor{VarColor}{\mathbf{M}} \in \textcolor{VarColor}{\mathcal{M}_{\text{ask}}}$.
        \State Partition the columns of $\textcolor{VarColor}{\mathbf{W}}$ into batches of size $\textcolor{VarColor}{B}$.
        \For{\textbf{each batch of columns} $\{\textcolor{VarColor}{\mathbf{w}_j}\}_{j=1}^B$}
            \For{\textbf{each column} $\textcolor{VarColor}{\mathbf{w}_j}$ in the batch \textbf{in parallel}}
                \State $\textcolor{VarColor}{\mathcal{S}_j} \gets \{ i \mid \textcolor{VarColor}{\mathbf{M}_{i,j}} = 0 \}$ \algorithmiccomment{Indices of pruned entries}
                \State Define QP:
                \begin{equation*}
                \begin{aligned}
                    &\min_{\textcolor{VarColor}{\Delta \mathbf{w}_j}} \textcolor{VarColor}{\Delta \mathbf{w}_j^T} \textcolor{VarColor}{\mathbf{H}_{\mathbf{W}}} \textcolor{VarColor}{\Delta \mathbf{w}_j} \\
                    &\text{s.t. } (\textcolor{VarColor}{\Delta \mathbf{w}_j})_i = -(\textcolor{VarColor}{\mathbf{w}_j})_i, \; \forall i \in \textcolor{VarColor}{\mathcal{S}_j}
                \end{aligned}
                \end{equation*}
            \EndFor
            \State $\textcolor{VarColor}{\{\Delta \mathbf{w}_j\}_{j=1}^B} \gets \textcolor{VarColor}{\texttt{QPSolve}}(\textcolor{VarColor}{\mathbf{H}_{\mathbf{W}}}, \{\textcolor{VarColor}{\mathbf{w}_j}\}_{j=1}^B, \{\textcolor{VarColor}{\mathcal{S}_j}\}_{j=1}^B)$ \algorithmiccomment{Batched rAPDHG solve with the shared Hessian}
            \State Update weights: $\textcolor{VarColor}{\mathbf{w}_j} \gets \textcolor{VarColor}{\mathbf{w}_j} + \textcolor{VarColor}{\Delta \mathbf{w}_j}, \quad \forall j$
        \EndFor
        \State If fewer than $95\%$ of the QPs in any batch converged, replace the update by the masked original weights; keep the candidate only if its reconstruction error is below that of the mask selector's weights.
    \EndFor
    \State $\textcolor{VarColor}{\mathbf{X}} \gets \textcolor{VarColor}{L}(\textcolor{VarColor}{\mathbf{X}})$ \algorithmiccomment{Recompute activations with the pruned and updated layer}
\EndFor

\vspace{0.5em}
\Statex \textbf{\textcolor{KeywordColor}{Return:}} Updated model $\textcolor{VarColor}{\hat{\mathcal{M}}}$.
\end{algorithmic}
\end{scriptsize}
\end{algorithm}

\niparagraph{Equality constraints.} Directly encoding the constraints from \autoref{eq:constraints} into the standard quadratic objective leads to a prohibitively large matrix of equalities, even though these constraints merely fix individual variables to constant values. We instead enforce them through variable bounds: $(\Delta \mathbf{w}_j)_i$ is boxed between $-(\mathbf{w}_j)_i$ and $-(\mathbf{w}_j)_i + 10^{-4}$, a slack that keeps the box non-degenerate for the solver, and the pruned entries are set to exactly zero after the solve, without the overhead of explicit equality matrices.

\niparagraph{Batching QP problems.} The QPs for all columns of a weight matrix may not fit in the memory of a single GPU, so we solve them in batches, which bounds memory while still exploiting the parallelism of many simultaneous QPs and lets \ours prune large LLMs on a single GPU.

\niparagraph{Hessian calculation.} Each weight matrix $\mathbf{W}$ has its own Hessian, estimated as the covariance of the \emph{input} activations to that matrix, i.e., the tensor entering the matrix multiplication with $\mathbf{W}$, captured by a hook on every weight matrix during a single calibration pass. For $b$ sequences of length $s$ and input dimension $d$, stacking all tokens into $\hat{\mathbf{X}}_{\mathbf{W}} \in \mathbb{R}^{bs \times d}$ and evaluating $\hat{\mathbf{X}}_{\mathbf{W}}^T \hat{\mathbf{X}}_{\mathbf{W}}$ gives $\mathbf{H}$ in \autoref{eq:constrained_objective} up to the scale factor $bs$, but storing all activations in accelerator memory is prohibitive for large $b$ and $s$. We therefore accumulate the covariance incrementally over sequences: with $\mathbf{x}_i \in \mathbb{R}^{s \times d}$ the input of sequence $i$, $\mathbf{H}_{\mathbf{W}} = \frac{1}{bs}\sum_{i=1}^b{\mathbf{x}_i^T \mathbf{x}_i}$, the token-mean covariance, yields the same minimizer without materializing $\hat{\mathbf{X}}_{\mathbf{W}}$; the solver tolerances refer to this normalized problem. Layers are processed sequentially: once the weight matrices of layer $L$ have been pruned and updated, the activations are recomputed with the updated weights (line 20 of \autoref{alg:layerwise_pruning}), so the masks and Hessians of layer $L+1$ reflect all earlier pruning decisions, the same protocol as SparseGPT and Thanos.

\niparagraph{Fallback for non-converged weight matrices.} If fewer than 95\% of the column QPs in any batch of a weight matrix terminate with an optimality certificate within the iteration budget, the QP update of that matrix is discarded and the candidate becomes the original weights with the masked entries zeroed. The candidate, whether the QP solution or these zeroed weights, replaces the mask selector's own update only if its reconstruction error is lower; otherwise the matrix keeps the selector's weights. \appref{app:implementation} reports how often this happens.

\section{Experiments}
\label{sec:results}

\niparagraph{Model, datasets, and evaluation.} We evaluate \ours on LLaMA 3.1, LLaMA 3.2 \citep{llama3}, Gemma 2 \citep{gemma2}, and Gemma 3 \citep{gemma3} family of models. Model accuracy is assessed on a range of zero-shot downstream tasks, including MMLU \citep{mmlu}, PIQA \citep{piqa}, ARC-Easy, ARC-Challenge \citep{arc}, WinoGrande \citep{winogrande}, and OpenBookQA \citep{openbookqa}, all of which are commonly used to evaluate LLM compression \citep{slim, wanda}. For zero-shot evaluations, we utilize the Language Model Evaluation Harness \citep{lm_eval} framework. In line with prior work \citep{wanda, sparsegpt, slim}, we also report the perplexity of the models on a language modeling task on the WikiText2 \citep{wikitext2} dataset.
\begin{figure}[t]
    \centering
    \includegraphics[width=0.55\linewidth]{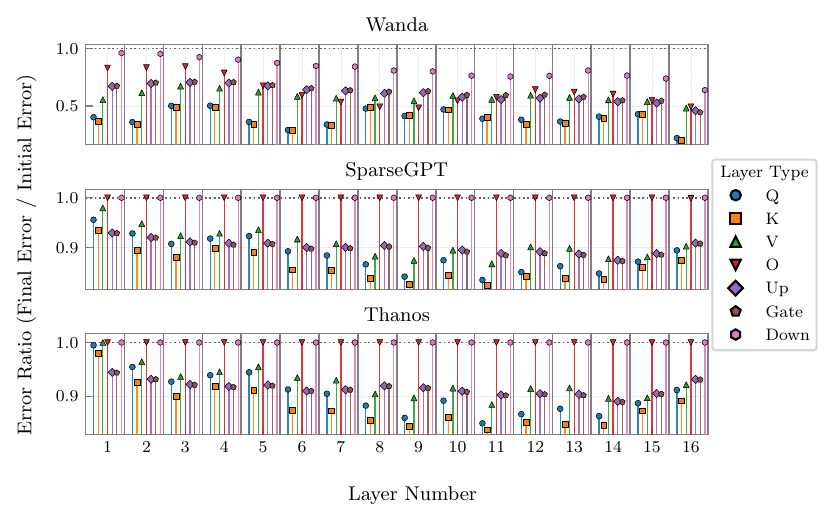}
    \caption{Ratio of \ours's layer-wise output error to that of the Wanda, SparseGPT, and Thanos updates on LLaMA 3.2 1B; values below 1 mean \ours lowers the error, and values near 1 mark weight matrices where the fallback kept the selector's update.}
    \label{fig:layerwise_error}
\end{figure}

\niparagraph{Baselines.} We compare \ours against one-shot pruning methods: Wanda \citep{wanda}, SparseGPT \citep{sparsegpt}, and Thanos \citep{thanos}. \appref{app:admm} adds the ADMM-based reconstruction of \citet{boza_admm}, which solves the same fixed-mask problem, and the 2:4 experiments add ProxSparse \citep{proxsparse}, which supports only 2:4 sparsity.
Additional details about the hyperparameters used in \ours are provided in \appref{app:implementation}, and the sensitivity of \ours to the calibration dataset size is reported in \appref{app:calibration}. \ours changes only the values of the retained weights, so an \ours-updated model inherits the latency, throughput, and memory profile of the baseline-pruned model; reported gains include: 2:4 sparsity gives $1.36\times$ to $1.57\times$ end-to-end throughput on LLaMA-2 7B and 13B with memory at 73\% of dense \citep{maskllm} and $1.24\times$ end-to-end on LLaMA-7B with sparse CUTLASS kernels \citep{wanda}, and 50\% and 60\% unstructured sparsity give $1.82\times$ and $2.16\times$ end-to-end CPU speedups on OPT-2.7B \citep{sparsegpt}; \ours's benefit is accuracy at unchanged inference cost.

\niparagraph{Model quality.} We evaluate the accuracy of \ours and the baseline pruning methods across 2:4 and unstructured sparsity. Wanda selects a mask without updating the remaining weights, whereas SparseGPT and Thanos update the weight values in addition to searching for the best mask. We couple the \ours weight update with the masks generated by each of these methods and compare the resulting models.

\autoref{tab:50_unstructured} summarizes the performance metrics for Wanda, SparseGPT, and Thanos with and without the \ours update mechanism for 50\% unstructured sparsity. Models pruned with the \ours weight update outperform the baseline weight updates in 12 of the 15 (model, mask selector) pairs in \autoref{tab:50_unstructured} and tie in one (LLaMA 3.1 8B with SparseGPT masks), providing up to 1.80 points of average accuracy improvement across six downstream tasks (Gemma 3 1B), with a mean improvement of 0.61 points across the 15 pairs.

\begin{table*}[tbp]
\centering
\setlength{\tabcolsep}{2pt}
\resizebox{\textwidth}{!}{
\begin{tabular}{l l l c c c c c c c | c}
\toprule
\multirow{2}{*}{\parbox[t]{1.5cm}{\centering Model}} &
\multirow{2}{*}{\parbox[t]{1cm}{\centering Mask\\Selection}} &
\multirow{2}{*}{\parbox[t]{1cm}{\centering Weight\\Update}} &
\multirow{2}{*}{Perplexity} &
\multicolumn{7}{c}{Metrics (\%)} \\
\cmidrule{5-11}
 & & & &
   \parbox[t]{\columndistance}{\centering \textcolor{blue!70!black}{MMLU}} &
   \parbox[t]{\columndistance}{\centering \textcolor{blue!70!black}{PIQA}} &
   \parbox[t]{\columndistance}{\centering \textcolor{blue!70!black}{Arc-E}} &
   \parbox[t]{\columndistance}{\centering \textcolor{blue!70!black}{Arc-C}} &
   \parbox[t]{\columndistance}{\centering \textcolor{blue!70!black}{Wino}} &
   \parbox[t]{\columndistance}{\centering \textcolor{blue!70!black}{OpenQA}} &
   \parbox[t]{\columndistance}{\centering \textbf{\textcolor{teal}{Avg.}}} \\
\midrule
\multirow{7}{*}{\parbox[t]{1.5cm}{\centering LLaMA\\3.1 8B}} & Dense & -- &  5.84 & 63.57 & 80.09 & 81.44 & 51.37 & 73.48 & 33.40 & 63.89 \\
 & Wanda & -- &  9.64 & 47.79 & 75.68 & 72.56 & 40.70 & 70.09 & 27.40 & 55.70 \\
 & \cellcolor{shade}{Wanda} & \cellcolor{shade}{\ours} & \cellcolor{shade}{9.37} & \cellcolor{shade}{\textbf{48.85}} & \cellcolor{shade}{\textbf{76.71}} & \cellcolor{shade}{\textbf{73.82}} & \cellcolor{shade}{\textbf{42.32}} & \cellcolor{shade}{\textbf{70.32}} & \cellcolor{shade}{\textbf{28.20}} & \cellcolor{shade}{\textbf{56.70}} \\
 & SparseGPT & SparseGPT &  9.30 & \textbf{51.32} & 76.17 & 73.02 & 41.30 & \textbf{70.88} & \textbf{29.40} & \textbf{57.02} \\
 & \cellcolor{shade}{SparseGPT} & \cellcolor{shade}{\ours} & \cellcolor{shade}{9.33} & \cellcolor{shade}{49.31} & \cellcolor{shade}{\textbf{76.61}} & \cellcolor{shade}{\textbf{74.28}} & \cellcolor{shade}{\textbf{\textbf{42.83}}} & \cellcolor{shade}{\textbf{70.88}} & \cellcolor{shade}{28.20} & \cellcolor{shade}{\textbf{57.02}} \\
 & Thanos & Thanos & 9.27 & \textbf{50.36} & \textbf{77.04} & \textbf{74.92} & \textbf{42.58} & \textbf{70.96} & \textbf{30.00} & \textbf{57.64} \\
 & \cellcolor{shade}{Thanos} & \cellcolor{shade}{\ours} & \cellcolor{shade}{9.35} & \cellcolor{shade}{50.17} & \cellcolor{shade}{76.50} & \cellcolor{shade}{74.16} & \cellcolor{shade}{41.89} & \cellcolor{shade}{70.24} & \cellcolor{shade}{28.40} & \cellcolor{shade}{56.89} \\
 \doublemidrule
\multirow{7}{*}{\parbox[t]{1.5cm}{\centering LLaMA\\3.2 1B}} & Dense & -- & 9.75 & 36.92 & 74.27 & 65.53 & 31.31 & 60.30 & 26.20 & 49.09
\\
 & Wanda & -- & 23.51 & 26.35 & 65.18 & 52.10 & 23.81 & 54.62 & 18.00 & 40.01 \\
 & \cellcolor{shade}{Wanda} & \cellcolor{shade}{\ours} & \cellcolor{shade}{18.84} & \cellcolor{shade}{\textbf{27.69}} & \cellcolor{shade}{\textbf{67.08}} & \cellcolor{shade}{\textbf{52.61}} & \cellcolor{shade}{\textbf{24.74}} & \cellcolor{shade}{\textbf{55.64}} & \cellcolor{shade}{\textbf{20.20}} & \cellcolor{shade}{\textbf{41.33}} \\
 & SparseGPT & SparseGPT & 18.84 & 25.71 & 67.85 & 54.29 & \textbf{26.54} & \textbf{57.70} & 22.00 & 42.35 \\
 & \cellcolor{shade}{SparseGPT} & \cellcolor{shade}{\ours} & \cellcolor{shade}{18.09} & \cellcolor{shade}{\textbf{26.95}} & \cellcolor{shade}{\textbf{68.01}} & \cellcolor{shade}{\textbf{54.59}} & \cellcolor{shade}{25.85} & \cellcolor{shade}{56.91} & \cellcolor{shade}{\textbf{24.00}} & \cellcolor{shade}{\textbf{42.72}} \\
 & Thanos & Thanos & 19.70 & 25.37 & 67.63 & 52.99 & \textbf{27.13} & 54.38 & \textbf{22.20} & 41.62 \\
 & \cellcolor{shade}{Thanos} & \cellcolor{shade}{\ours} & \cellcolor{shade}{18.77} & \cellcolor{shade}{\textbf{25.99}} & \cellcolor{shade}{\textbf{68.23}} & \cellcolor{shade}{\textbf{53.49}} & \cellcolor{shade}{26.45} & \cellcolor{shade}{\textbf{55.88}} & \cellcolor{shade}{21.60} & \cellcolor{shade}{\textbf{41.94}} \\
\doublemidrule
\multirow{7}{*}{\parbox[t]{1.5cm}{\centering LLaMA\\3.2 3B}} & Dense & -- & 7.81 & 54.13 & 76.55 & 74.28 & 42.75 & 69.38 & 30.60 & 57.95 \\
 & Wanda & -- & 12.92 & 40.79 & 72.03 & 65.45 & 32.34 & 63.69 & 25.40 & 49.95 \\
 & \cellcolor{shade}{Wanda} & \cellcolor{shade}{\ours} & \cellcolor{shade}{12.24} & \cellcolor{shade}{\textbf{43.11}} & \cellcolor{shade}{\textbf{72.47}} & \cellcolor{shade}{\textbf{66.50}} & \cellcolor{shade}{\textbf{33.53}} & \cellcolor{shade}{\textbf{66.38}} & \cellcolor{shade}{\textbf{26.20}} & \cellcolor{shade}{\textbf{51.37}} \\
& SparseGPT & SparseGPT & 12.32 & 37.96 & \textbf{73.45} & 65.19 & 33.02 & 66.38 & 25.20 & 50.20 \\
 & \cellcolor{shade}{SparseGPT} & \cellcolor{shade}{\ours} & \cellcolor{shade}{12.43} & \cellcolor{shade}{\textbf{40.54}} & \cellcolor{shade}{\textbf{73.45}} & \cellcolor{shade}{\textbf{66.37}} & \cellcolor{shade}{\textbf{35.07}} & \cellcolor{shade}{\textbf{66.69}} & \cellcolor{shade}{\textbf{26.20}} & \cellcolor{shade}{\textbf{51.39}} \\
& Thanos & Thanos & 12.26 & 40.11 & 72.80 & 64.77 & 32.85 & \textbf{67.72} & 26.60 & 50.81 \\
 & \cellcolor{shade}{Thanos} & \cellcolor{shade}{\ours} & \cellcolor{shade}{12.40} & \cellcolor{shade}{\textbf{41.51}} & \cellcolor{shade}{\textbf{73.23}} & \cellcolor{shade}{\textbf{65.07}} & \cellcolor{shade}{\textbf{34.39}} & \cellcolor{shade}{67.25} & \cellcolor{shade}{\textbf{27.00}} & \cellcolor{shade}{\textbf{51.41}} \\
\doublemidrule
\multirow{7}{*}{\parbox[t]{1.5cm}{\centering Gemma\\3 1B}} & Dense & -- & 14.17 & 24.95 & 74.81 & 71.93 & 35.41 & 58.72 & 28.80 & 49.10 \\
 & Wanda & -- & {32.96} & 22.97 & 67.19 & 61.03 & 26.37 & 55.72 & 20.00 & 42.21 \\
 & \cellcolor{shade}{Wanda} & \cellcolor{shade}{\ours} & \cellcolor{shade}{28.90} & \cellcolor{shade}{\textbf{23.96}} & \cellcolor{shade}{\textbf{69.48}} & \cellcolor{shade}{\textbf{62.84}} & \cellcolor{shade}{\textbf{28.58}} & \cellcolor{shade}{\textbf{56.83}} & \cellcolor{shade}{\textbf{22.40}} & \cellcolor{shade}{\textbf{44.01}} \\
 & SparseGPT & SparseGPT & {28.34} & 24.85 & 68.88 & \textbf{60.94} & 26.62 & 55.49 & 21.40 & 43.03 \\
 & \cellcolor{shade}{SparseGPT} & \cellcolor{shade}{\ours} & \cellcolor{shade}{27.35} & \cellcolor{shade}{\textbf{25.73}} & \cellcolor{shade}{\textbf{69.75}} & \cellcolor{shade}{60.90} & \cellcolor{shade}{\textbf{27.82}} & \cellcolor{shade}{\textbf{56.35}} & \cellcolor{shade}{\textbf{22.00}} & \cellcolor{shade}{\textbf{43.76}} \\
 & Thanos & Thanos & {28.65} & 23.09 & \textbf{69.75} & 62.16 & \textbf{27.99} & \textbf{56.51} & \textbf{23.80} & 43.88 \\
 & \cellcolor{shade}{Thanos} & \cellcolor{shade}{\ours} & \cellcolor{shade}{28.14} & \cellcolor{shade}{\textbf{24.70}} & \cellcolor{shade}{69.64} & \cellcolor{shade}{\textbf{63.43}} & \cellcolor{shade}{27.39} & \cellcolor{shade}{55.96} & \cellcolor{shade}{23.20} & \cellcolor{shade}{\textbf{44.05}} \\
\doublemidrule
\multirow{7}{*}{\parbox[t]{1.5cm}{\centering Gemma\\2 2B}} & Dense & -- & 68.69 & 49.33 & 78.24 & 80.22 & 46.93 & 68.82 & 31.40 & 59.16 \\
 & Wanda & -- & {327.45} & 34.17 & \textbf{74.16} & 69.78 & \textbf{34.30} & \textbf{62.83} & \textbf{26.40} & \textbf{50.27} \\
 & \cellcolor{shade}{Wanda} & \cellcolor{shade}{\ours} & \cellcolor{shade}{215.63} & \cellcolor{shade}{\textbf{34.86}} & \cellcolor{shade}{73.99} & \cellcolor{shade}{\textbf{71.38}} & \cellcolor{shade}{32.59} & \cellcolor{shade}{61.96} & \cellcolor{shade}{25.80} & \cellcolor{shade}{50.10} \\
 & SparseGPT & SparseGPT & 234.68 & 35.59 & 73.61 & 69.99 & 34.22 & \textbf{65.82} & \textbf{28.20} & 51.24 \\
 & \cellcolor{shade}{SparseGPT} & \cellcolor{shade}{\ours} & \cellcolor{shade}{{241.09}} & \cellcolor{shade}{\textbf{37.59}} & \cellcolor{shade}{\textbf{73.83}} & \cellcolor{shade}{\textbf{70.62}} & \cellcolor{shade}{\textbf{35.07}} & \cellcolor{shade}{64.72} & \cellcolor{shade}{27.80} & \cellcolor{shade}{\textbf{51.60}} \\
 & Thanos & Thanos & {276.97} & 30.62 & 73.18 & 67.72 & 33.62 & 63.22 & \textbf{26.80} & 49.19 \\
 & \cellcolor{shade}{Thanos} & \cellcolor{shade}{\ours} & \cellcolor{shade}{250.15} & \cellcolor{shade}{\textbf{32.72}} & \cellcolor{shade}{\textbf{73.72}} & \cellcolor{shade}{\textbf{68.81}} & \cellcolor{shade}{\textbf{34.13}} & \cellcolor{shade}{\textbf{63.85}} & \cellcolor{shade}{26.40} & \cellcolor{shade}{\textbf{49.94}} \\
\bottomrule
\end{tabular}
}
\caption{WikiText2 perplexity and zero-shot accuracy at 50\% unstructured sparsity. \ours improves the average accuracy in 12 of 15 (model, mask selector) pairs and ties one. Within each block, the better of the two Weight Update rows is bolded.}
\label{tab:50_unstructured}
\vspace{-15pt}
\end{table*}

\autoref{tab:2_4} presents the results of pruning transformer models using 2:4 semi-structured sparsity. In these experiments, we prune only the weight matrices in the multilayer perceptron (MLP) components, leaving the self-attention layers dense, which yields an overall sparsity of 38\% to 41\%. Roughly 80\% of the prunable parameters in these families reside in the MLP blocks, whereas attention projections are parameter-light yet accuracy-critical: in preliminary experiments, extending 2:4 to attention at the 1B to 8B scale collapsed every method, with or without \ours, to near-random accuracy, and a benchmark at chance level orders methods by noise rather than merit (even MaskLLM, which learns 2:4 masks with thousands of GPU-hours, keeps the most sensitive layers dense \citep{maskllm}). This setup also follows the common deployment pattern where 2:4 sparsity is applied only to the compute-dominant MLP weights, for which NVIDIA's sparse tensor cores provide hardware acceleration. \ours is orthogonal to where 2:4 is applied, so stronger full-2:4 selectors compose with it unchanged. Under 2:4 sparsity, \ours improves the six-task average in 17 of 19 pairs (mean $+0.56$ points, up to $+3.21$ on LLaMA 3.2 1B with Wanda masks).

\begin{table*}[tbp]
\centering
\setlength{\tabcolsep}{2pt}
\resizebox{\textwidth}{!}{
\begin{tabular}{l l l c c c c c c c | c}
\toprule
\multirow{2}{*}{\parbox[t]{1.5cm}{\centering Model}} &
\multirow{2}{*}{\parbox[t]{1cm}{\centering Mask\\Selection}} &
\multirow{2}{*}{\parbox[t]{1cm}{\centering Weight\\Update}} &
\multirow{2}{*}{Perplexity} &
\multicolumn{7}{c}{Metrics (\%)} \\
\cmidrule{5-11}
 & & & &
   \parbox[t]{\columndistance}{\centering \textcolor{blue!70!black}{MMLU}} &
   \parbox[t]{\columndistance}{\centering \textcolor{blue!70!black}{PIQA}} &
   \parbox[t]{\columndistance}{\centering \textcolor{blue!70!black}{Arc-E}} &
   \parbox[t]{\columndistance}{\centering \textcolor{blue!70!black}{Arc-C}} &
   \parbox[t]{\columndistance}{\centering \textcolor{blue!70!black}{Wino}} &
   \parbox[t]{\columndistance}{\centering \textcolor{blue!70!black}{OpenQA}} &
   \parbox[t]{\columndistance}{\centering \textbf{\textcolor{teal}{Avg.}}} \\
\midrule
\multirow{7}{*}{\parbox[t]{1.5cm}{\centering LLaMA\\3.1 8B}} & Dense & -- & 5.84 & 63.57 & 80.09 & 81.44 & 51.37 & 73.48 & 33.40 & 63.89 \\
 & Wanda & -- & {13.54} & 43.42 & 73.18 & 69.23 & 35.32 & 67.32 & \textbf{25.80} & 52.38 \\
 & \cellcolor{shade}Wanda & \cellcolor{shade}\ours & \cellcolor{shade}12.58 & \cellcolor{shade}{\textbf{45.45}} & \cellcolor{shade}{\textbf{73.39}} & \cellcolor{shade}{\textbf{69.57}} & \cellcolor{shade}{\textbf{36.18}} & \cellcolor{shade}{\textbf{68.90}} & \cellcolor{shade}{25.20} & \cellcolor{shade}{\textbf{53.12}} \\
 & SparseGPT & SparseGPT & 12.37 & 45.62 & \textbf{73.83} & 69.15 & 35.84 & 69.22 & 25.60 & 53.21 \\
 & \cellcolor{shade}SparseGPT & \cellcolor{shade}\ours & \cellcolor{shade}{{12.54}} & \cellcolor{shade}\textbf{46.04} & \cellcolor{shade}{73.72} & \cellcolor{shade}{\textbf{69.95}} & \cellcolor{shade}\textbf{36.77} & \cellcolor{shade}\textbf{69.61} & \cellcolor{shade}\textbf{27.00} & \cellcolor{shade}\textbf{53.85} \\
 & Thanos & Thanos & 12.66 & 44.39 & 73.94 & 69.57 & 36.18 & \textbf{68.90} & 25.20 & 53.03 \\
 & \cellcolor{shade}Thanos & \cellcolor{shade}\ours & \cellcolor{shade}{{12.80}} & \cellcolor{shade}{\textbf{44.41}} & \cellcolor{shade}\textbf{74.05} & \cellcolor{shade}\textbf{69.95} & \cellcolor{shade}{\textbf{36.43}} & \cellcolor{shade}{68.59} & \cellcolor{shade}{\textbf{25.60}} & \cellcolor{shade}{\textbf{53.17}} \\
 \doublemidrule
\multirow{9}{*}{\parbox[t]{1.5cm}{\centering LLaMA\\3.2 1B}} & Dense & -- & 9.75 & 36.92 & 74.27 & 65.53 & 31.31 & 60.30 & 26.20 & 49.09 \\
 & Wanda & -- & 30.43 & 23.32 & 63.55 & 47.56 & \textbf{23.63} & 55.25 & 15.00 & 38.05 \\
 & \cellcolor{shade}Wanda & \cellcolor{shade}\ours & \cellcolor{shade}{{48.23}} & \cellcolor{shade}\textbf{24.80} & \cellcolor{shade}\textbf{66.10} & \cellcolor{shade}\textbf{58.04} & \cellcolor{shade}{23.55} & \cellcolor{shade}{55.25} & \cellcolor{shade}\textbf{19.80} & \cellcolor{shade}\textbf{41.26} \\%
 & SparseGPT & SparseGPT & {21.98} & 23.05 & 65.45 & 52.15 & 25.17 & \textbf{57.62} & 17.60 & 40.17 \\
 & \cellcolor{shade}SparseGPT & \cellcolor{shade}\ours & \cellcolor{shade}{21.40} & \cellcolor{shade}{\textbf{23.40}} & \cellcolor{shade}{\textbf{65.72}} & \cellcolor{shade}{\textbf{52.78}} & \cellcolor{shade}\textbf{25.51} & \cellcolor{shade}{57.06} & \cellcolor{shade}{\textbf{18.60}} & \cellcolor{shade}{\textbf{40.51}} \\%
 & Thanos & Thanos & {22.80} & \textbf{24.09} & \textbf{65.67} & 51.68 & \textbf{25.00} & 52.96 & \textbf{17.60} & 39.50 \\
 & \cellcolor{shade}Thanos & \cellcolor{shade}\ours & \cellcolor{shade}{22.26} & \cellcolor{shade}{23.41} & \cellcolor{shade}{65.34} & \cellcolor{shade}{\textbf{52.23}} & \cellcolor{shade}{23.72} & \cellcolor{shade}{\textbf{55.96}} & \cellcolor{shade}{16.80} & \cellcolor{shade}{\textbf{39.58}} \\
 & ProxSparse & -- & 41.95 & \textbf{23.64} & 61.21 & 42.38 & \textbf{22.53} & 53.67 & 16.00 & 36.57 \\
 & \cellcolor{shade}ProxSparse & \cellcolor{shade}\ours & \cellcolor{shade}28.53 & \cellcolor{shade}23.07 & \cellcolor{shade}\textbf{63.38} & \cellcolor{shade}\textbf{47.90} & \cellcolor{shade}\textbf{22.53} & \cellcolor{shade}\textbf{54.78} & \cellcolor{shade}\textbf{16.40} & \cellcolor{shade}\textbf{38.01} \\
\doublemidrule
\multirow{9}{*}{\parbox[t]{1.5cm}{\centering LLaMA\\3.2 3B}} & Dense & -- & 7.81 & 54.13 & 76.55 & 74.28 & 42.75 & 69.38 & 30.60 & 57.95 \\
 & Wanda & -- & {18.51} & 34.30 & 70.73 & 60.69 & 30.72 & 61.17 & \textbf{24.80} & 47.07 \\
 & \cellcolor{shade}Wanda & \cellcolor{shade}\ours & \cellcolor{shade}16.64 & \cellcolor{shade}{\textbf{37.15}} & \cellcolor{shade}{\textbf{70.78}} & \cellcolor{shade}{\textbf{61.95}} & \cellcolor{shade}{\textbf{31.14}} & \cellcolor{shade}{\textbf{62.51}} & \cellcolor{shade}24.60 & \cellcolor{shade}{\textbf{48.02}} \\
 & SparseGPT & SparseGPT & 16.19 & 36.13 & 70.29 & 63.01 & 30.46 & \textbf{64.72} & 25.00 & 48.27 \\
 & \cellcolor{shade}SparseGPT & \cellcolor{shade}\ours & \cellcolor{shade}{{16.36}} & \cellcolor{shade}\textbf{38.03} & \cellcolor{shade}\textbf{70.84} & \cellcolor{shade}\textbf{63.17} & \cellcolor{shade}\textbf{32.17} & \cellcolor{shade}{63.69} & \cellcolor{shade}\textbf{25.60} & \cellcolor{shade}\textbf{48.92} \\
 & Thanos & Thanos & 16.24 & 35.55 & 70.35 & 61.28 & 29.78 & \textbf{63.30} & 24.20 & 47.41 \\
 & \cellcolor{shade}Thanos & \cellcolor{shade}\ours & \cellcolor{shade}{{16.49}} & \cellcolor{shade}{\textbf{35.72}} & \cellcolor{shade}{\textbf{70.62}} & \cellcolor{shade}{\textbf{62.04}} & \cellcolor{shade}{\textbf{30.97}} & \cellcolor{shade}{63.22} & \cellcolor{shade}\textbf{25.60} & \cellcolor{shade}{\textbf{48.03}} \\
 & ProxSparse & -- & 19.50 & 24.66 & 68.12 & 56.31 & 27.82 & 58.56 & 20.00 & 42.58 \\
 & \cellcolor{shade}ProxSparse & \cellcolor{shade}\ours & \cellcolor{shade}18.28 & \cellcolor{shade}\textbf{31.76} & \cellcolor{shade}\textbf{69.53} & \cellcolor{shade}\textbf{60.27} & \cellcolor{shade}\textbf{28.84} & \cellcolor{shade}\textbf{60.30} & \cellcolor{shade}\textbf{20.60} & \cellcolor{shade}\textbf{45.22} \\
\doublemidrule
\multirow{9}{*}{\parbox[t]{1.5cm}{\centering Gemma\\3 1B}} & Dense & -- & 14.17 & 24.95 & 74.81 & 71.93 & 35.41 & 58.72 & 28.80 & 49.10 \\
 & Wanda & -- & {60.74} & \textbf{23.74} & \textbf{65.51} & \textbf{56.78} & 22.35 & 52.72 & \textbf{19.80} & \textbf{40.15} \\
 & \cellcolor{shade}Wanda & \cellcolor{shade}\ours & \cellcolor{shade}23.25 & \cellcolor{shade}23.25 & \cellcolor{shade}63.38 & \cellcolor{shade}51.14 & \cellcolor{shade}\textbf{24.06} & \cellcolor{shade}{\textbf{54.30}} & \cellcolor{shade}18.20 & \cellcolor{shade}39.06 \\
 & SparseGPT & SparseGPT & {44.87} & 24.83 & \textbf{66.76} & 57.70 & 23.29 & \textbf{55.96} & 19.40 & 41.32 \\
 & \cellcolor{shade}SparseGPT & \cellcolor{shade}\ours & \cellcolor{shade}{42.66} & \cellcolor{shade}{\textbf{25.11}} & \cellcolor{shade}{66.27} & \cellcolor{shade}{\textbf{58.96}} & \cellcolor{shade}{\textbf{23.89}} & \cellcolor{shade}{55.80} & \cellcolor{shade}{\textbf{20.60}} & \cellcolor{shade}\textbf{41.77} \\
 & Thanos & Thanos & {48.50} & 25.23 & 65.89 & \textbf{59.30} & 23.12 & 53.59 & \textbf{20.80} & 41.32 \\
 & \cellcolor{shade}Thanos & \cellcolor{shade}\ours & \cellcolor{shade}{44.91} & \cellcolor{shade}\textbf{25.83} & \cellcolor{shade}{\textbf{66.00}} & \cellcolor{shade}{58.63} & \cellcolor{shade}{\textbf{23.29}} & \cellcolor{shade}{\textbf{54.70}} & \cellcolor{shade}{20.00} & \cellcolor{shade}{\textbf{41.41}} \\
 & ProxSparse & -- & 41.02 & 23.01 & \textbf{66.00} & \textbf{54.34} & 22.44 & \textbf{55.88} & \textbf{20.20} & \textbf{40.31} \\
 & \cellcolor{shade}ProxSparse & \cellcolor{shade}\ours & \cellcolor{shade}52.99 & \cellcolor{shade}\textbf{24.13} & \cellcolor{shade}64.74 & \cellcolor{shade}53.70 & \cellcolor{shade}\textbf{22.61} & \cellcolor{shade}52.25 & \cellcolor{shade}17.00 & \cellcolor{shade}39.07 \\
\doublemidrule
\multirow{9}{*}{\parbox[t]{1.5cm}{\centering Gemma\\2 2B}} & Dense & -- & 68.69 & 49.33 & 78.24 & 80.22 & 46.93 & 68.82 & 31.40 & 59.16 \\
 & Wanda & -- & 421.01 & 34.34 & 71.33 & 68.10 & 30.97 & 61.40 & \textbf{26.40} & 48.76 \\
 & \cellcolor{shade}Wanda & \cellcolor{shade}\ours & \cellcolor{shade}229.69 & \cellcolor{shade}{\textbf{34.44}} & \cellcolor{shade}{\textbf{71.87}} & \cellcolor{shade}\textbf{68.90} & \cellcolor{shade}{\textbf{33.87}} & \cellcolor{shade}{\textbf{62.27}} & \cellcolor{shade}{25.00} & \cellcolor{shade}{\textbf{49.39}} \\
 & SparseGPT & SparseGPT & 251.71 & \textbf{32.84} & 71.76 & \textbf{68.73} & \textbf{32.42} & 61.88 & 23.40 & 48.51 \\
 & \cellcolor{shade}SparseGPT & \cellcolor{shade}\ours & \cellcolor{shade}{227.99} & \cellcolor{shade}{32.77} & \cellcolor{shade}{71.76} & \cellcolor{shade}{67.47} & \cellcolor{shade}{32.17} & \cellcolor{shade}\textbf{63.38} & \cellcolor{shade}{\textbf{24.40}} & \cellcolor{shade}{\textbf{48.66}} \\
 & Thanos & Thanos & 256.58 & 31.02 & 70.73 & \textbf{67.72} & 32.08 & \textbf{62.51} & 24.80 & 48.14 \\
 & \cellcolor{shade}Thanos & \cellcolor{shade}\ours & \cellcolor{shade}{239.20} & \cellcolor{shade}{\textbf{32.58}} & \cellcolor{shade}{\textbf{71.16}} & \cellcolor{shade}{67.47} & \cellcolor{shade}{\textbf{32.25}} & \cellcolor{shade}{60.85} & \cellcolor{shade}{\textbf{25.20}} & \cellcolor{shade}{\textbf{48.25}} \\
 & ProxSparse & -- & 176.03 & 37.19 & \textbf{71.98} & 67.55 & \textbf{34.47} & 61.48 & \textbf{25.00} & 49.61 \\
 & \cellcolor{shade}ProxSparse & \cellcolor{shade}\ours & \cellcolor{shade}254.03 & \cellcolor{shade}\textbf{38.27} & \cellcolor{shade}71.27 & \cellcolor{shade}\textbf{68.60} & \cellcolor{shade}33.53 & \cellcolor{shade}\textbf{61.88} & \cellcolor{shade}24.60 & \cellcolor{shade}\textbf{49.69} \\
\bottomrule
\end{tabular}
}
\caption{WikiText2 perplexity and zero-shot accuracy at 2:4 sparsity applied to the MLP layers only (38\% to 41\% end-to-end sparsity). \ours improves the average accuracy in 17 of 19 (model, mask selector) pairs. Within each block, the better of the two Weight Update rows is bolded.}
\label{tab:2_4}
\vspace{-10pt}
\end{table*}

\niparagraph{Perplexity versus downstream accuracy.} For some (model, mask selector) pairs, perplexity and average downstream accuracy move in opposite directions: the row with better WikiText2 perplexity is not always the row with higher zero-shot accuracy. The two metrics diverge in both directions: on LLaMA 3.2 1B with Wanda masks at 2:4, perplexity rises from 30.43 to 48.23 while accuracy rises 3.21 points, and Gemma 3 1B with Wanda masks at 2:4 shows the reverse (perplexity 60.74 to 23.25, accuracy $-1.09$). The pair therefore measures different things rather than exposing a failure mode of \ours. Gemma 2 2B's dense perplexity is already 68.69, so its large sparse perplexities are family-specific. For compressed LLMs specifically, \citet{accuracy_not_all_you_need} show that perplexity averages log-probabilities, so per-token degradations can be cancelled elsewhere, and conclude that perplexity alone is inadequate for evaluating compression; see also \citet{same_pretraining_loss, scaling_laws_unreliable}. Because the fallback keeps the layer-wise error at or below the selector's, these mismatches are cases where lower layer-wise error does not help end to end (\autoref{sec:limitations}). We therefore do not treat perplexity as the primary measure of pruning quality. We retain the perplexity columns in the tables for comparability with prior work, but encourage the reader to give more weight to the downstream-accuracy columns when judging \ours against the baselines.

\niparagraph{Higher sparsity ratios.} \ours's gains are largest at 60\% unstructured sparsity: 11 of 15 (model, mask selector) pairs improve, by 1.01 points on average and up to 3.97 points on LLaMA 3.2 3B (\appref{app:60unstr_full}). Pruning more weights leaves more reconstruction error for \ours to absorb, which translates into both lower perplexity and higher zero-shot accuracy than Wanda on its own, and improves SparseGPT and Thanos masks over their own updates in 6 of 10 pairs. On LLaMA 3.2 3B, for instance, \ours raises Wanda's average accuracy from 38.77\% to 42.74\%.

\niparagraph{Where \ours helps.} \ours helps most where the mask selector leaves the most reconstruction error. At 50\% unstructured sparsity the improvements are small or within experimental noise, whereas at 60\% they become larger on average (mean $+1.01$ versus $+0.61$ points). The method also improves Wanda masks the most, whose ``no weight update'' baseline leaves the most reconstruction error unabsorbed and therefore the most headroom for a principled column-wise QP update to recover accuracy. When the mask selector already performs its own weight update (SparseGPT, Thanos), the additional benefit from \ours is smaller and occasionally regresses, which suggests that the two update schemes can interact in ways that are not uniformly constructive. At 2:4 MLP-only sparsity the results are more sensitive to the specific combination of model family and mask selector. Overall, \ours improves 40 of the 49 (model, mask selector, sparsity) cells and ties one. The eight regressions range from $-0.03$ to $-1.24$ points against largest gains of $+3.21$ and $+3.97$, and are spread over models and selectors: LLaMA 3.1 8B with Thanos masks ($-0.75$) and Gemma 2 2B with Wanda masks ($-0.17$) at 50\%, Gemma 3 1B with Wanda ($-1.09$) and ProxSparse ($-1.24$) masks at 2:4, and four cells with SparseGPT or Thanos masks between $-0.03$ and $-0.38$ at 60\%. Thanos couples mask selection and weight update in one greedy procedure, whereas \ours is a pure weight-update step on any fixed mask; on the very same Thanos mask, \ours's update helps in 12 of 15 settings across the three sparsity regimes. The same headroom mechanism explains the fallback statistics in \appref{app:implementation}.

\niparagraph{Generalization to the Qwen 2.5 family.} Results carry over to Qwen 2.5 \citep{qwen2.5} at 0.5B to 14B parameters for unstructured sparsity and are mixed at 2:4: \ours improves 11 of 15 (model size, mask selector) pairs at 50\% (mean $+0.37$ points) and 11 of 15 at 60\% (mean $+0.81$, up to $+4.19$ for Wanda on Qwen 2.5 3B), whereas at 2:4 (MLP-only) it improves 8 of 9 pairs at the 0.5B to 3B scales but regresses for all three selectors on Qwen 2.5 7B (mean over all 15 pairs $-0.22$). Full tables are in \appref{app:qwen}.

\niparagraph{Comparison with alternative optimizers.} ADAM \citep{adam} does not substitute for the constrained QP solver: ADAM occasionally achieves competitive results on larger models, but it often converges to suboptimal solutions and can even diverge on smaller models, whereas rAPDHG certifies convergence per column and the fallback keeps the selector's update where it does not converge (\appref{app:adam}). \appref{app:finetuning} reports a data-matched recovery fine-tuning experiment: ADAM on the same 128 calibration samples on top of Wanda or SparseGPT masks for 1 to 128 iterations recovers at most $+0.17$ average-accuracy points before overfitting collapses the model, and never approaches \ours.

\niparagraph{Layer-wise error improvement.} \ours lowers the layer-wise error of LLaMA 3.2 1B wherever the solver converges (\autoref{fig:layerwise_error}). The figure shows the output error ratio of all pruned weight matrices, defined as $\frac{MSE(Y_\text{\our}, Y_\text{dense})}{MSE(Y_\text{other}, Y_\text{dense})}$, where $MSE$ denotes the mean squared error across the calibration dataset. The ratios of about 1.0 for the attention output and MLP down projections under SparseGPT and Thanos masks mark weight matrices where the fallback of \autoref{sec:methodology} kept the selector's own update.

\niparagraph{Pruning time analysis.} On a single NVIDIA H100 with 80 GB of memory, \ours prunes LLaMA 3.2 1B and Gemma 3 1B in about \SI{2.5}{\hour} each, Gemma 2 2B in \SI{5.5}{\hour}, LLaMA 3.2 3B in \SI{7.0}{\hour}, and LLaMA 3.1 8B in \SI{40.0}{\hour} with 60 GB of peak GPU memory.
Pruning time grows with model size and is not noticeably affected by architecture: models of similar size take comparable time to prune. Mask selection alone takes 15 to 50 seconds with Wanda and 65 to 308 seconds with SparseGPT on these models, so \ours is orders of magnitude slower than these baselines and than ADMM-based reconstruction; pruning is a one-time offline cost, and the single-GPU times are conservative since the independent column QPs can be dispatched across devices, a headroom we have not measured. LLaMA 3.1 70B with Wanda masks takes about 350 hours on one H100 and improves from 67.04\% to 67.30\% at 50\% sparsity and from 60.74\% to 61.37\% at 60\% (\appref{app:70b}).

\section{Related work}
\label{sec:related_work}

Pruning removes weights that contribute little to a network's outputs; \appref{app:related_work} gives an extended discussion, and we summarize the two main lines here. Layer-wise methods prune one layer at a time under the assumption that keeping each layer's output close to the dense model's keeps the end-to-end error small. They build on the Optimal Brain Surgeon (OBS) framework \citep{obs} applied per layer \citep{obc}, whose full second-order machinery is too expensive at LLM scale; Optimal BERT Surgeon \citep{obs_bert} scales OBS-style approximations to transformers via block-diagonal Hessians, SparseGPT \citep{sparsegpt} and Thanos \citep{thanos} solve the reconstruction approximately, DSnoT \citep{dsnot} adjusts the mask without weight training, and Wanda \citep{wanda} drops the weight update entirely. \ours fits in this line: it keeps the selector's mask and replaces the heuristic update with a principled, parallel QP solve. Exact solvers of the same post-mask convex problem exist: \citet{boza_admm} use ADMM, and ALPS \citep{alps} refines its ADMM-selected support with a vectorized GPU conjugate-gradient solve on the shared Hessian; FISTAPruner \citep{fistapruner} and CHITA \citep{chita} instead relax or constrain the mask jointly with the weights. On ADMM's own masks \ours matches or slightly exceeds its update, by at most 0.15 accuracy points, while ADMM is faster (about 14 minutes for LLaMA-7B in the original paper; \appref{app:admm}); \ours differs in terminating on a certified per-column tolerance and composing with any mask selector. Wanda++ \citep{wandapp} also composes a selector with a block-level update. End-to-end methods use global information instead: Optimal Brain Damage \citep{obd} and the network-wide OBS rely on second-order statistics, WoodFisher \citep{woodfisher} and LLM Surgeon \citep{llmsurgeon} approximate them with structured curvature, SparseLLM \citep{sparsellm} and Elsa \citep{elsa} prune globally with decomposed or ADMM subproblems, and MaskLLM \citep{maskllm} and ProxSparse \citep{proxsparse} learn 2:4 masks with optimization loops heavier than a single pass over calibration data. Quantization \citep{quantization_survey, quantization_survey2}, low-rank adaptation \citep{slim, lqlora, slope}, and knowledge distillation \citep{knowledge_distillation} target other sources of redundancy and can be layered on top of pruning.

\section{Conclusion}
\ours{} reformulates post-training weight reconstruction as batched, column-wise QPs that share one Hessian per weight matrix.
This yields per-column \emph{optimal} updates for the reconstruction objective given the estimated Hessian. A batched primal-dual solver computes them with a certificate per column, and the step drops into common mask selectors (Wanda, SparseGPT, Thanos).
In our experiments on a single NVIDIA H100, \ours{} improves average zero-shot accuracy in 40 of 49 (model, mask selector, sparsity) settings on LLaMA and Gemma, by up to 3.97 points, and matches ADMM-based exact reconstruction on its own masks.
The gains are largest at 60\% sparsity and require no post-pruning fine-tuning.
Together, these results deliver a principled and scalable approach to accurate one-shot post-training pruning.

\subsection{Limitations}
\label{sec:limitations}

\ours has several limitations that constrain where it is most useful and point to directions for future work. First, on runtime cost, \ours is substantially slower than SparseGPT and Wanda on a per-layer basis, because it runs an iterative first-order QP solver to convergence rather than applying a closed-form or single-sweep update. This cost is the price of the joint column optimization, and for pipelines where pruning time dominates, SparseGPT or ADMM-based reconstruction remains the faster choice. Second, on the objective itself, \ours minimizes layer-wise reconstruction error, which is not perfectly aligned with end-to-end task performance; in some settings, a lower reconstruction error does not translate into higher downstream accuracy, a known limitation of layer-wise pruning that \ours inherits. Third, on solver tolerance, our reported results use a relative and absolute rAPDHG tolerance of 0.01; tightening it to $10^{-3}$ or $10^{-4}$ changes the six-task average by at most about 0.6 points with no consistent direction (\appref{app:tolerance}), so the default loses no accuracy. Fourth, with SparseGPT and Thanos masks the fallback triggers on 28\% to 29\% of the weight matrices of LLaMA 3.2 1B, the one setting where we measured it, and at 2:4 sparsity \ours regresses on Qwen 2.5 7B for all three selectors ($-2.84$ to $-3.29$ points).

\acks{We extend our gratitude towards James Laudon and Karolina Dziugaite for reviewing the paper and providing insightful feedback. We also thank the extended team at Google DeepMind who enabled and supported this research direction.
This work was also supported in part by NSERC Discovery Grants (RGPIN-06516, DGECR00303), the Canada Research Chairs program, the Ontario Early Researcher Award, the Digital Research Alliance of Canada (\url{www.alliancecan.ca}), and a Google unrestricted gift (JAX AI Stack Research Award).
}

\ifshort
\niparagraph{Appendix.} The appendix (implementation details, derivations, the ADMM comparison, full per-task tables, ablations, LLaMA-3.1-70B results, and extended related work) is in the full version at \url{https://arxiv.org/abs/2512.13886} and in the supplementary material.
\fi

\FloatBarrier
\ifshort\else\clearpage\fi

\bibliography{acml25}

@inproceedings{obs,
	title        = {Optimal brain surgeon: Extensions and performance comparisons},
	author       = {Hassibi, Babak and Stork, David and Wolff, Gregory},
	year         = 1993,
	booktitle    = {Advances in Neural Information Processing Systems}
}

@inproceedings{obd,
	title        = {Optimal brain damage},
	author       = {LeCun, Yann and Denker, John and Solla, Sara},
	year         = 1990,
	booktitle    = {Advances in Neural Information Processing Systems}
}

@inproceedings{wanda,
	title        = {A simple and effective pruning approach for large language models},
	author       = {Sun, Mingjie and Liu, Zhuang and Bair, Anna and Kolter, J Zico},
	year         = 2024,
	booktitle    = {International Conference on Learning Representations (ICLR)}
}

@inproceedings{sparsegpt,
	title        = {{SparseGPT}: Massive language models can be accurately pruned in one-shot},
	author       = {Frantar, Elias and Alistarh, Dan},
	year         = 2023,
	booktitle    = {Proceedings of the International Conference on Machine Learning (ICML)}
}

@inproceedings{lqlora,
	title        = {{LQ-LoRA: Low-rank Plus Quantized Matrix Decomposition for Efficient Language Model Finetuning}},
	author       = {Guo, Han and Greengard, Philip and Xing, Eric P and Kim, Yoon},
	year         = 2024,
	booktitle    = {International Conference on Learning Representations (ICLR)}
}

@inproceedings{mkor,
	title        = {{MKOR: Momentum-Enabled Kronecker-Factor-Based Optimizer Using Rank-1 Updates}},
	author       = {Mozaffari, Mohammad and Li, Sikan and Zhang, Zhao and Mehri Dehnavi, Maryam},
	year         = 2023,
	booktitle    = {Advances in Neural Information Processing Systems (NeurIPS)}
}

@inproceedings{woodfisher,
	title        = {{WoodFisher}: Efficient second-order approximation for neural network compression},
	author       = {Singh, Sidak Pal and Alistarh, Dan},
	year         = 2020,
	booktitle    = {Advances in Neural Information Processing Systems (NeurIPS)}
}

@article{opt,
	title        = {{OPT}: Open pre-trained transformer language models},
	author       = {Zhang, Susan and Roller, Stephen and Goyal, Naman and Artetxe, Mikel and others},
	year         = 2022,
	journal      = {arXiv preprint arXiv:2205.01068}
}

@article{llama2,
	title        = {{Llama} 2: Open foundation and fine-tuned chat models},
	author       = {Touvron, Hugo and Martin, Louis and Stone, Kevin and Albert, Peter and others},
	year         = 2023,
	journal      = {arXiv preprint arXiv:2307.09288}
}

@inproceedings{mmlu,
	title        = {Measuring massive multitask language understanding},
	author       = {Hendrycks, Dan and Burns, Collin and Basart, Steven and Zou, Andy and others},
	year         = 2021,
	booktitle    = {International Conference on Learning Representations (ICLR)}
}

@inproceedings{piqa,
	title        = {{PIQA}: Reasoning about physical commonsense in natural language},
	author       = {Bisk, Yonatan and Zellers, Rowan and Gao, Jianfeng and Choi, Yejin and others},
	year         = 2020,
	booktitle    = {Proceedings of the AAAI Conference on Artificial Intelligence}
}

@article{arc,
	title        = {Think you have solved question answering? try {ARC}, the {AI2} reasoning challenge},
	author       = {Clark, Peter and Cowhey, Isaac and Etzioni, Oren and Khot, Tushar and others},
	year         = 2018,
	journal      = {arXiv preprint arXiv:1803.05457}
}

@article{winogrande,
	title        = {{WinoGrande}: An adversarial {Winograd} schema challenge at scale},
	author       = {Sakaguchi, Keisuke and Bras, Ronan Le and Bhagavatula, Chandra and Choi, Yejin},
	year         = 2021,
	journal      = {Communications of the ACM},
	publisher    = {ACM New York, NY, USA},
	volume       = 64,
	number       = 9,
	pages        = {99--106}
}

@inproceedings{openbookqa,
	title        = {Can a suit of armor conduct electricity? a new dataset for open book question answering},
	author       = {Mihaylov, Todor and Clark, Peter and Khot, Tushar and Sabharwal, Ashish},
	year         = 2018,
	booktitle    = {Proceedings of the 2018 Conference on Empirical Methods in Natural Language Processing (EMNLP)}
}

@misc{lm_eval,
	title        = {A framework for few-shot language model evaluation},
	author       = {Gao, Leo and Tow, Jonathan and Abbasi, Baber and Biderman, Stella and others},
	year         = 2024,
	month        = jul,
	publisher    = {Zenodo},
	doi          = {10.5281/zenodo.12608602},
	url          = {https://zenodo.org/records/12608602},
	version      = {v0.4.3}
}

@inproceedings{wikitext2,
	title        = {Pointer Sentinel Mixture Models},
	author       = {Stephen Merity and Caiming Xiong and James Bradbury and Richard Socher},
	year         = 2017,
	booktitle    = {International Conference on Learning Representations (ICLR)}
}

@article{c4,
	title        = {Exploring the Limits of Transfer Learning with a Unified Text-to-Text Transformer},
	author       = {Colin Raffel and Noam Shazeer and Adam Roberts and Katherine Lee and others},
	year         = 2020,
	journal      = {Journal of Machine Learning Research},
	archiveprefix = {arXiv},
	eprint       = {1910.10683},
	volume       = 21,
	number       = 140,
	pages        = {1--67}
}

@inproceedings{adamw,
	title        = {Decoupled weight decay regularization},
	author       = {Loshchilov, Ilya and Hutter, Frank},
	year         = 2019,
	booktitle    = {International Conference on Learning Representations (ICLR)}
}

@incollection{quantization_survey,
	title        = {A survey of quantization methods for efficient neural network inference},
	author       = {Gholami, Amir and Kim, Sehoon and Dong, Zhen and Yao, Zhewei and others},
	year         = 2022,
	booktitle    = {Low-Power Computer Vision},
	publisher    = {Chapman and Hall/CRC}
}

@article{quantization_survey2,
	title        = {A comprehensive survey on model quantization for deep neural networks in image classification},
	author       = {Rokh, Babak and Azarpeyvand, Ali and Khanteymoori, Alireza},
	year         = 2023,
	journal      = {ACM Transactions on Intelligent Systems and Technology},
	publisher    = {ACM New York, NY},
	volume       = 14,
	number       = 6,
	pages        = {1--50}
}

@inproceedings{obc,
	title        = {Optimal brain compression: A framework for accurate post-training quantization and pruning},
	author       = {Frantar, Elias and Alistarh, Dan},
	year         = 2022,
	booktitle    = {Advances in Neural Information Processing Systems (NeurIPS)}
}

@article{llama3,
	title        = {The {Llama} 3 herd of models},
	author       = {Dubey, Abhimanyu and Jauhri, Abhinav and Pandey, Abhinav and Kadian, Abhishek and others},
	year         = 2024,
	journal      = {arXiv preprint arXiv:2407.21783}
}

@article{gemini,
	title        = {{Gemini} 1.5: Unlocking multimodal understanding across millions of tokens of context},
	author       = {{Gemini Team} and Georgiev, Petko and Lei, Ving Ian and Burnell, Ryan and others},
	year         = 2024,
	journal      = {arXiv preprint arXiv:2403.05530}
}

@inproceedings{maskllm,
	title        = {{MaskLLM}: Learnable semi-structured sparsity for large language models},
	author       = {Fang, Gongfan and Yin, Hongxu and Muralidharan, Saurav and Heinrich, Greg and others},
	year         = 2024,
	booktitle    = {Advances in Neural Information Processing Systems (NeurIPS)}
}

@article{gemma2,
	title        = {{Gemma} 2: Improving open language models at a practical size},
	author       = {{Gemma Team} and Riviere, Morgane and Pathak, Shreya and Sessa, Pier Giuseppe and others},
	year         = 2024,
	journal      = {arXiv preprint arXiv:2408.00118}
}

@article{thanos,
	title        = {Thanos: A Block-wise Pruning Algorithm for Efficient Large Language Model Compression},
	author       = {Ilin, Ivan and Richtarik, Peter},
	year         = 2025,
	journal      = {arXiv preprint arXiv:2504.05346}
}

@inproceedings{slim,
	title        = {{SLiM: One-shot Quantized Sparse Plus Low-rank Approximation of LLMs}},
	author       = {Mozaffari, Mohammad and Yazdanbakhsh, Amir and Mehri Dehnavi, Maryam},
	year         = 2025,
	booktitle    = {Proceedings of the International Conference on Machine Learning (ICML)},
	url          = {https://openreview.net/forum?id=4UfRP8MopP}
}

@inproceedings{slope,
	title        = {{SLoPe: Double-Pruned Sparse Plus Lazy Low-Rank Adapter Pretraining of LLMs}},
	author       = {Mozaffari, Mohammad and Yazdanbakhsh, Amir and Zhang, Zhao and Mehri Dehnavi, Maryam},
	year         = 2025,
	booktitle    = {International Conference on Learning Representations (ICLR)}
}

@article{gemma3,
	title        = {{Gemma} 3 technical report},
	author       = {{Gemma Team} and Kamath, Aishwarya and Ferret, Johan and Pathak, Shreya and others},
	year         = 2025,
	journal      = {arXiv preprint arXiv:2503.19786}
}

@inproceedings{adam,
	title        = {{Adam}: A method for stochastic optimization},
	author       = {Kingma, Diederik P and Ba, Jimmy},
	year         = 2015,
	booktitle    = {International Conference on Learning Representations (ICLR)}
}

@article{gemini2.5,
	title        = {Gemini 2.5: Pushing the frontier with advanced reasoning, multimodality, long context, and next generation agentic capabilities},
	author       = {Comanici, Gheorghe and Bieber, Eric and Schaekermann, Mike and Pasupat, Ice and others},
	year         = 2025,
	journal      = {arXiv preprint arXiv:2507.06261}
}

@article{knowledge_distillation,
	title        = {Knowledge distillation: A survey},
	author       = {Gou, Jianping and Yu, Baosheng and Maybank, Stephen J and Tao, Dacheng},
	year         = 2021,
	journal      = {International Journal of Computer Vision},
	publisher    = {Springer},
	volume       = 129,
	number       = 6,
	pages        = {1789--1819}
}

@article{sparse_ft,
	title        = {Enabling high-sparsity foundational llama models with efficient pretraining and deployment},
	author       = {Agarwalla, Abhinav and Gupta, Abhay and Marques, Alexandre and Pandit, Shubhra and others},
	year         = 2024,
	journal      = {arXiv preprint arXiv:2405.03594}
}

@article{pdqp,
	title        = {A practical and optimal first-order method for large-scale convex quadratic programming},
	author       = {Lu, Haihao and Yang, Jinwen},
	year         = 2026,
	journal      = {Mathematical Programming},
	volume       = 215,
	pages        = {771--808},
	doi          = {10.1007/s10107-025-02241-0}
}

@article{mpax,
	title        = {{MPAX}: Mathematical Programming in {JAX}},
	author       = {Lu, Haihao and Peng, Zedong and Yang, Jinwen},
	year         = 2024,
	journal      = {arXiv preprint arXiv:2412.09734}
}

@inproceedings{proxsparse,
	title        = {ProxSparse: Regularized Learning of Semi-Structured Sparsity Masks for Pretrained LLMs},
	author       = {Liu, Hongyi and Saha, Rajarshi and Jia, Zhen and Park, Youngsuk and others},
	year         = 2025,
	booktitle    = {Proceedings of the International Conference on Machine Learning (ICML)}
}

@article{deepseekr1,
	title        = {{DeepSeek-R1} incentivizes reasoning in {LLMs} through reinforcement learning},
	author       = {Guo, Daya and Yang, Dejian and Zhang, Haowei and Song, Junxiao and others},
	year         = 2025,
	journal      = {Nature},
	volume       = 645,
	pages        = {633--638}
}

@article{sparsity_survery,
	title        = {Sparsity in deep learning: Pruning and growth for efficient inference and training in neural networks},
	author       = {Hoefler, Torsten and Alistarh, Dan and Ben-Nun, Tal and Dryden, Nikoli and others},
	year         = 2021,
	journal      = {Journal of Machine Learning Research},
	volume       = 22,
	number       = 241,
	pages        = {1--124}
}

@article{qwen2.5,
	title        = {{Qwen2.5} Technical Report},
	author       = {Jinze Bai and Shuai Bai and Yunfei Chu and Zeyu Cui and others},
	year         = 2024,
	journal      = {arXiv preprint arXiv:2409.11586}
}

@inproceedings{dsnot,
	title        = {Dynamic Sparse No Training: Training-Free Fine-tuning for Sparse {LLM}s},
	author       = {Zhang, Yuxin and Zhao, Lirui and Lin, Mingbao and Sun, Yunyun and Yao, Yiwu and Han, Xingjia and Tanner, Jason and Liu, Shiwei and Ji, Rongrong},
	year         = 2024,
	booktitle    = {International Conference on Learning Representations (ICLR)}
}

@inproceedings{obs_bert,
	title        = {The Optimal {BERT} Surgeon: Scalable and Accurate Second-Order Pruning for Large Language Models},
	author       = {Kurtic, Eldar and Campos, Daniel and Nguyen, Tuan and Frantar, Elias and Kurtz, Mark and Fineran, Benjamin and Goin, Michael and Alistarh, Dan},
	year         = 2022,
	booktitle    = {Proceedings of the 2022 Conference on Empirical Methods in Natural Language Processing (EMNLP)}
}

@article{boza_admm,
	title        = {Fast and Effective Weight Update for Pruned Large Language Models},
	author       = {Bo{\v{z}}a, Vladim{\'\i}r},
	year         = 2024,
	journal      = {Transactions on Machine Learning Research},
	url          = {https://openreview.net/forum?id=1hcpXd9Jir}
}

@inproceedings{accuracy_not_all_you_need,
	title        = {Accuracy is Not All You Need},
	author       = {Dutta, Abhinav and Krishnan, Sanjeev and Kwatra, Nipun and Ramjee, Ramachandran},
	year         = 2024,
	booktitle    = {Advances in Neural Information Processing Systems (NeurIPS)}
}

@inproceedings{same_pretraining_loss,
	title        = {Same Pre-training Loss, Better Downstream: Implicit Bias Matters for Language Models},
	author       = {Liu, Hong and Xie, Sang Michael and Li, Zhiyuan and Ma, Tengyu},
	year         = 2023,
	booktitle    = {Proceedings of the International Conference on Machine Learning (ICML)}
}

@inproceedings{scaling_laws_unreliable,
	title        = {Scaling Laws Are Unreliable for Downstream Tasks: A Reality Check},
	author       = {Lourie, Nicholas and Hu, Michael Y. and Cho, Kyunghyun},
	year         = 2025,
	booktitle    = {Findings of the Association for Computational Linguistics: EMNLP 2025}
}

@inproceedings{alps,
	title        = {{ALPS}: Improved Optimization for Highly Sparse One-Shot Pruning for Large Language Models},
	author       = {Meng, Xiang and Behdin, Kayhan and Wang, Haoyue and Mazumder, Rahul},
	year         = 2024,
	booktitle    = {Advances in Neural Information Processing Systems (NeurIPS)}
}

@inproceedings{fistapruner,
	title        = {{FISTAPruner}: Layer-wise Post-training Pruning for Large Language Models},
	author       = {Zhao, Pengxiang and Hu, Hanyu and Li, Ping and Zheng, Yi and Wang, Zhefeng and Yuan, Xiaoming},
	year         = 2025,
	booktitle    = {Proceedings of the 2025 Conference on Empirical Methods in Natural Language Processing (EMNLP)}
}

@inproceedings{sparsellm,
	title        = {{SparseLLM}: Towards Global Pruning of Pre-trained Language Models},
	author       = {Bai, Guangji and Li, Yijiang and Ling, Chen and Kim, Kibaek and Zhao, Liang},
	year         = 2024,
	booktitle    = {Advances in Neural Information Processing Systems (NeurIPS)}
}

@inproceedings{llmsurgeon,
	title        = {The {LLM} Surgeon},
	author       = {van der Ouderaa, Tycho F. A. and Nagel, Markus and van Baalen, Mart and Asano, Yuki M. and Blankevoort, Tijmen},
	year         = 2024,
	booktitle    = {International Conference on Learning Representations (ICLR)}
}

@inproceedings{wandapp,
	title        = {{Wanda++}: Pruning Large Language Models via Regional Gradients},
	author       = {Yang, Yifan and Zhen, Kai and Ganesh, Bhavana and Galstyan, Aram and others},
	year         = 2025,
	booktitle    = {Findings of the Association for Computational Linguistics: ACL 2025}
}

@inproceedings{elsa,
	title        = {The Unseen Frontier: Pushing the Limits of {LLM} Sparsity with Surrogate-Free {ADMM}},
	author       = {Lee, Kwanhee and Jang, Hyeondo and Lee, Dongyeop and Alistarh, Dan and Lee, Namhoon},
	year         = 2026,
	booktitle    = {International Conference on Learning Representations (ICLR)}
}

@inproceedings{chita,
	title        = {Fast as {CHITA}: Neural Network Pruning with Combinatorial Optimization},
	author       = {Benbaki, Riade and Chen, Wenyu and Meng, Xiang and Hazimeh, Hussein and Ponomareva, Natalia and Zhao, Zhe and Mazumder, Rahul},
	year         = 2023,
	booktitle    = {Proceedings of the International Conference on Machine Learning (ICML)}
}

\ifshort\else
\clearpage
\appendix

\begingroup
{\LARGE\bfseries Appendix\par}
\vspace{0.8em}
{\large\bfseries Table of Contents\par}
\vspace{0.3em}
\hrule height 0.8pt
\vspace{0.3em}
\startcontents[appendix]
\printcontents[appendix]{}{1}{\setcounter{tocdepth}{2}}
\vspace{0.3em}
\hrule height 0.8pt
\endgroup

\section{Implementation details and hyperparameters}
\label{app:implementation}

Following previous work \citep{sparsegpt, wanda, slim, thanos}, we use 128 samples, each with 2048 tokens from the C4 dataset \citep{c4} for calibration.

We set the relative and absolute tolerance of the rAPDHG QP solver in MPAX to 0.01 and the maximum number of iterations to 100,000; both refer to the token-mean normalized Hessian of \autoref{sec:methodology}, and the QPs are solved in fp32 with the solution cast back to the model dtype. The column QPs of a weight matrix with input dimension $d$ are solved in batches of $B = \min(\text{columns}, \lfloor 1.5 \cdot 2^{30} / d^2 \rfloor)$ (384 columns at $d = 2048$, 96 at $d = 4096$). If fewer than 95\% of the QPs in any batch terminate with an optimality certificate within the iteration budget, the QP update of the whole matrix is discarded and replaced by the original weights with the masked entries zeroed; the resulting candidate, or the QP solution otherwise, is applied only if its reconstruction error is lower than that of the mask selector's own weights, which the matrix keeps otherwise. \autoref{tab:hyperparameters} summarizes the key hyperparameters employed in our method.

\niparagraph{Fallback frequency.} Instrumented pruning of LLaMA 3.2 1B shows that the fallback triggers for 0\% of weight matrices with Wanda masks, 28\% with SparseGPT masks, and 29\% with Thanos masks, and exclusively for the attention output and MLP down projections; it never triggers for the Q, K, V, Up, and Gate projections. The error ratios of about 1.0 for exactly these projections in \autoref{fig:layerwise_error} are the reverted matrices. The fallback costs compute but guarantees that the layer-wise reconstruction error is never worse than with the mask selector's own update.

For all other baselines used in our work, we either use their publicly available checkpoint or use their repositories to reproduce their results with their default hyperparameters.

\begin{table}[bt]
\centering
\caption{Key hyperparameters of \ours and of the ADAM baseline of \autoref{app:adam} (last two rows).}
\label{tab:hyperparameters}
\begin{tabular}{lc}
\toprule
\textbf{Hyperparameter} & \textbf{Value} \\
\midrule
Calibration Samples & 128 \\
Tokens per Sample & 2048 \\
Dataset for Calibration & C4 \\
Relative Tolerance (rAPDHG) & 0.01 \\
Absolute Tolerance (rAPDHG) & 0.01 \\
Maximum Iterations (rAPDHG) & 100,000 \\
ADAM Learning Rate & \{$10^{-2}, 10^{-3}, 10^{-4}, 10^{-5}$\} \\
ADAM Weight Decay & 0 \\
\bottomrule
\end{tabular}
\end{table}

\FloatBarrier
\section{Reformulation as an unconstrained quadratic program}
\label{app:unconstrained}

As an alternative to the constrained formulation used in \autoref{sec:qp} (\autoref{eq:constrained_objective}), we can reformulate each column-wise problem as an unconstrained quadratic program. This can be useful in settings where solvers are optimized for unconstrained problems or when eliminating constraints enables more efficient optimization. Although our implementation adopts the constrained approach for reasons discussed below, we include the unconstrained version for completeness.

The key idea is to eliminate the equality constraints in \autoref{eq:constraints} by substituting them directly into the objective. For a given column $j$, define $\mathcal{I}_j$ as the set of indices where the mask is one (i.e., unpruned weights), and let $\mathcal{S}_j$ denote the complement set (i.e., pruned weights, where the mask is zero).

We reorder the entries of the change vector $\Delta\mathbf{w}_j$ and the shared Hessian matrix $\mathbf{H} = \mathbf{X}_l^T \mathbf{X}_l$ based on this partitioning, as shown in \autoref{eq:block_partitioning}.

\begin{equation}
\Delta\mathbf{w}_j = \begin{bmatrix} \Delta\mathbf{w}_{\mathcal{I}_j} \\ \Delta\mathbf{w}_{\mathcal{S}_j} \end{bmatrix}, \quad
\mathbf{H} = \begin{bmatrix}
\mathbf{H}_{\mathcal{I}_j\mathcal{I}_j} & \mathbf{H}_{\mathcal{I}_j\mathcal{S}_j} \\
\mathbf{H}_{\mathcal{S}_j\mathcal{I}_j} & \mathbf{H}_{\mathcal{S}_j\mathcal{S}_j}
\end{bmatrix}
\label{eq:block_partitioning}
\end{equation}

As established in \autoref{eq:constraints}, the entries of $\Delta\mathbf{w}_j$ corresponding to $\mathcal{S}_j$ are fixed: $(\Delta\mathbf{w}_j)_i = -(\mathbf{w}_j)_i$ for all $i \in \mathcal{S}_j$. Substituting these fixed values into the quadratic objective yields the expanded form in \autoref{eq:expanded_objective}.

\begin{equation}
\begin{aligned}
\Delta\mathbf{w}_j^T \mathbf{H} \Delta\mathbf{w}_j &=
\Delta\mathbf{w}_{\mathcal{I}_j}^T \mathbf{H}_{\mathcal{I}_j\mathcal{I}_j} \Delta\mathbf{w}_{\mathcal{I}_j} +
2 \Delta\mathbf{w}_{\mathcal{I}_j}^T \mathbf{H}_{\mathcal{I}_j\mathcal{S}_j}  \\
\Delta\mathbf{w}_{\mathcal{S}_j} +
\Delta\mathbf{w}_{\mathcal{S}_j}^T \mathbf{H}_{\mathcal{S}_j\mathcal{S}_j} \Delta\mathbf{w}_{\mathcal{S}_j}
\end{aligned}
\label{eq:expanded_objective}
\end{equation}

Since $\Delta\mathbf{w}_{\mathcal{S}_j} = -\mathbf{w}_{\mathcal{S}_j}$, we substitute this to obtain the unconstrained objective in \autoref{eq:unconstrained_objective}.

\begin{equation}
\min_{\Delta\mathbf{w}_{\mathcal{I}_j}}
\Delta\mathbf{w}_{\mathcal{I}_j}^T \mathbf{H}_{\mathcal{I}_j\mathcal{I}_j} \Delta\mathbf{w}_{\mathcal{I}_j}
- 2 \Delta\mathbf{w}_{\mathcal{I}_j}^T \mathbf{H}_{\mathcal{I}_j\mathcal{S}_j} \mathbf{w}_{\mathcal{S}_j}
+ \mathbf{w}_{\mathcal{S}_j}^T \mathbf{H}_{\mathcal{S}_j\mathcal{S}_j} \mathbf{w}_{\mathcal{S}_j}
\label{eq:unconstrained_objective}
\end{equation}

The final term in \autoref{eq:unconstrained_objective} is constant with respect to the optimization variable $\Delta\mathbf{w}_{\mathcal{I}_j}$ and can therefore be omitted. This results in the unconstrained quadratic program in \autoref{eq:final_unconstrained_qp}.

\begin{equation}
\underset{\Delta\mathbf{w}_{\mathcal{I}_j}}{\text{minimize}} \quad
\Delta\mathbf{w}_{\mathcal{I}_j}^T \mathbf{Q}_j \Delta\mathbf{w}_{\mathcal{I}_j}
+ \mathbf{c}_j^T \Delta\mathbf{w}_{\mathcal{I}_j}
\label{eq:final_unconstrained_qp}
\end{equation}

where the problem-specific matrix and vector are defined as:
\begin{equation}
\mathbf{Q}_j = \mathbf{H}_{\mathcal{I}_j\mathcal{I}_j}, \quad
\mathbf{c}_j = -2 \mathbf{H}_{\mathcal{I}_j\mathcal{S}_j} \mathbf{w}_{\mathcal{S}_j}
\label{eq:unconstrained_qp_params}
\end{equation}

This formulation eliminates the need for explicit constraints, but introduces column-dependent variation in problem dimensions. Specifically, the size of $\mathbf{Q}_j$ and $\mathbf{c}_j$ varies with the number of unpruned weights in each column. Consequently, the unconstrained QPs have heterogeneous shapes and objectives across columns, making them more difficult to batch and parallelize efficiently on accelerators like GPUs or TPUs. This motivates our choice to adopt the constrained formulation in \autoref{sec:qp}, where the problem structure is uniform and well-suited for high-throughput parallel execution.

\FloatBarrier
\section{Matched-protocol comparison with ADMM-based reconstruction}
\label{app:admm}

The ADMM-based method of \citet{boza_admm} solves the same fixed-mask convex reconstruction problem as \ours. To isolate the weight-update step, ADMM selects the mask, and on that same mask, with the same calibration data and evaluation protocol, we compare three weight-update options: (i) none (original weights), (ii) ADMM's own update, and (iii) \ours. \autoref{tab:admm} reports the results at 50\% and 60\% unstructured sparsity.

\ours matches or slightly exceeds ADMM's update in all six average-accuracy and all six perplexity comparisons (never behind, strictly ahead in 11 of 12), but the margins are small (0.03 to 0.15 accuracy points), exactly as theory predicts. Both methods are exact solvers of the same convex fixed-mask problem: \citet{boza_admm} prove asymptotic convergence of the ADMM iterates to the optimum, and rAPDHG converges linearly to a per-column optimum with a primal-dual tolerance certificate at termination. We read the near-identical results as mutual validation rather than a deficiency of either method. This suggests that further fixed-mask solver gains are small at these scales, shifting attention toward mask selection and end-to-end objectives \citep{elsa}.

Within the subproblem, the two methods differ in mechanics, not solution quality. ADMM caches one ridge-regularized inverse and runs a fixed 20 iterations without a convergence check (its theoretical results do not bound the speed of convergence), whereas \ours terminates on a certified per-column tolerance, making non-convergence detectable rather than silent. On raw pruning time, ADMM is faster (about 14 minutes end-to-end for LLaMA-7B on two Quadro RTX 5000 GPUs in the original paper, versus 2.5 to 40 hours for \ours at the 1B to 8B scale); where pruning time dominates, it is the pragmatic choice. \ours's contribution is the certified, accelerator-batched route to the same optimum, a reference solution against which faster update heuristics can be validated, and modularity: ADMM is another mask selector \ours composes with, after Wanda, SparseGPT, Thanos, and ProxSparse.

\begin{table*}[tb]
\centering
\setlength{\tabcolsep}{2pt}
\resizebox{\textwidth}{!}{
\begin{tabular}{l l l c c c c c c c | c}
\toprule
\multirow{2}{*}{\parbox[t]{2.2cm}{\centering Sparsity}} &
\multirow{2}{*}{\parbox[t]{1.5cm}{\centering Model}} &
\multirow{2}{*}{\parbox[t]{1cm}{\centering Weight\\Update}} &
\multirow{2}{*}{Perplexity} &
\multicolumn{7}{c}{Metrics (\%)} \\
\cmidrule{5-11}
 & & & &
   \parbox[t]{\columndistance}{\centering \textcolor{blue!70!black}{MMLU}} &
   \parbox[t]{\columndistance}{\centering \textcolor{blue!70!black}{PIQA}} &
   \parbox[t]{\columndistance}{\centering \textcolor{blue!70!black}{Arc-E}} &
   \parbox[t]{\columndistance}{\centering \textcolor{blue!70!black}{Arc-C}} &
   \parbox[t]{\columndistance}{\centering \textcolor{blue!70!black}{Wino}} &
   \parbox[t]{\columndistance}{\centering \textcolor{blue!70!black}{OpenQA}} &
   \parbox[t]{\columndistance}{\centering \textbf{\textcolor{teal}{Avg.}}} \\
\midrule
\multirow{9}{*}{\parbox[t]{2.2cm}{\centering 50\%\\unstructured}} & \multirow{3}{*}{\parbox[t]{1.5cm}{\centering LLaMA\\3.2 1B}} & -- & 25.35 & 25.97 & 65.40 & 50.72 & 22.87 & 54.46 & 21.20 & 40.10 \\
 & & ADMM & \textbf{17.42} & 27.00 & 68.88 & 55.51 & 27.13 & 56.27 & 23.00 & 42.97 \\
 & & \cellcolor{shade}\ours & \cellcolor{shade}\textbf{17.42} & \cellcolor{shade}26.93 & \cellcolor{shade}69.59 & \cellcolor{shade}55.68 & \cellcolor{shade}27.13 & \cellcolor{shade}56.35 & \cellcolor{shade}22.40 & \cellcolor{shade}\textbf{43.01} \\
\cmidrule{2-11}
 & \multirow{3}{*}{\parbox[t]{1.5cm}{\centering LLaMA\\3.2 3B}} & -- & 14.03 & 38.37 & 72.74 & 65.24 & 34.13 & 63.06 & 25.40 & 49.82 \\
 & & ADMM & 11.58 & 43.19 & 73.94 & 67.42 & 34.47 & 66.69 & 28.60 & 52.38 \\
 & & \cellcolor{shade}\ours & \cellcolor{shade}\textbf{11.52} & \cellcolor{shade}42.57 & \cellcolor{shade}74.16 & \cellcolor{shade}66.92 & \cellcolor{shade}35.58 & \cellcolor{shade}66.85 & \cellcolor{shade}28.80 & \cellcolor{shade}\textbf{52.48} \\
\cmidrule{2-11}
 & \multirow{3}{*}{\parbox[t]{1.5cm}{\centering Gemma\\3 1B}} & -- & 29.85 & 23.69 & 69.10 & 62.08 & 27.22 & 56.35 & 22.40 & 43.47 \\
 & & ADMM & 26.63 & 24.05 & 70.08 & 63.80 & 27.73 & 56.51 & 25.20 & 44.56 \\
 & & \cellcolor{shade}\ours & \cellcolor{shade}\textbf{26.32} & \cellcolor{shade}24.71 & \cellcolor{shade}70.08 & \cellcolor{shade}63.64 & \cellcolor{shade}27.90 & \cellcolor{shade}56.75 & \cellcolor{shade}25.20 & \cellcolor{shade}\textbf{44.71} \\
\doublemidrule
\multirow{9}{*}{\parbox[t]{2.2cm}{\centering 60\%\\unstructured}} & \multirow{3}{*}{\parbox[t]{1.5cm}{\centering LLaMA\\3.2 1B}} & -- & 80.23 & 23.00 & 57.56 & 37.21 & 20.48 & 52.72 & 13.60 & 34.10 \\
 & & ADMM & 34.53 & 24.73 & 64.91 & 47.14 & 23.21 & 54.93 & 18.60 & 38.92 \\
 & & \cellcolor{shade}\ours & \cellcolor{shade}\textbf{34.13} & \cellcolor{shade}24.72 & \cellcolor{shade}64.85 & \cellcolor{shade}47.22 & \cellcolor{shade}23.29 & \cellcolor{shade}55.01 & \cellcolor{shade}18.60 & \cellcolor{shade}\textbf{38.95} \\
\cmidrule{2-11}
 & \multirow{3}{*}{\parbox[t]{1.5cm}{\centering LLaMA\\3.2 3B}} & -- & 36.68 & 27.91 & 65.45 & 52.36 & 24.91 & 57.06 & 17.60 & 40.88 \\
 & & ADMM & 19.02 & 32.84 & 69.37 & 57.74 & 27.99 & 60.77 & 22.80 & 45.25 \\
 & & \cellcolor{shade}\ours & \cellcolor{shade}\textbf{18.88} & \cellcolor{shade}33.12 & \cellcolor{shade}69.10 & \cellcolor{shade}58.25 & \cellcolor{shade}27.99 & \cellcolor{shade}61.01 & \cellcolor{shade}22.60 & \cellcolor{shade}\textbf{45.35} \\
\cmidrule{2-11}
 & \multirow{3}{*}{\parbox[t]{1.5cm}{\centering Gemma\\3 1B}} & -- & 68.02 & 24.04 & 63.71 & 52.36 & 20.73 & 53.20 & 19.20 & 38.87 \\
 & & ADMM & 50.55 & 25.17 & 65.29 & 55.89 & 22.70 & 53.83 & 19.60 & 40.41 \\
 & & \cellcolor{shade}\ours & \cellcolor{shade}\textbf{49.84} & \cellcolor{shade}25.39 & \cellcolor{shade}64.64 & \cellcolor{shade}55.68 & \cellcolor{shade}23.12 & \cellcolor{shade}54.14 & \cellcolor{shade}20.00 & \cellcolor{shade}\textbf{40.50} \\
\bottomrule
\end{tabular}
}
\caption{Matched-protocol comparison with ADMM-based reconstruction \citep{boza_admm}. ADMM selects the mask; on that same mask and the same calibration data, the rows compare no weight update (--), ADMM's own update, and \ours. Within each block, the better perplexity and average accuracy of the two updates are bolded.}
\label{tab:admm}
\end{table*}

\FloatBarrier
\section{Full results at 60\% unstructured sparsity}
\label{app:60unstr_full}

\autoref{tab:60_unstructured} reports the full per-task numbers at 60\% unstructured sparsity on the LLaMA-3 and Gemma families, summarized in \autoref{sec:results}. \ours improves the six-task average in 11 of the 15 (model, mask selector) pairs, with a mean gain of 1.01 points and gains up to 3.97 points on LLaMA-3.2-3B; the four regressions ($-0.03$ to $-0.38$ points) all occur with SparseGPT or Thanos masks.

\begin{table*}[tb]
\centering
\setlength{\tabcolsep}{2pt}
\resizebox{\textwidth}{!}{
\begin{tabular}{l l l c c c c c c c | c}
\toprule
\multirow{2}{*}{\parbox[t]{1.5cm}{\centering Model}} &
\multirow{2}{*}{\parbox[t]{1cm}{\centering Mask\\Selection}} &
\multirow{2}{*}{\parbox[t]{1cm}{\centering Weight\\Update}} &
\multirow{2}{*}{Perplexity} &
\multicolumn{7}{c}{Metrics (\%)} \\
\cmidrule{5-11}
 & & & &
   \parbox[t]{\columndistance}{\centering \textcolor{blue!70!black}{MMLU}} &
   \parbox[t]{\columndistance}{\centering \textcolor{blue!70!black}{PIQA}} &
   \parbox[t]{\columndistance}{\centering \textcolor{blue!70!black}{Arc-E}} &
   \parbox[t]{\columndistance}{\centering \textcolor{blue!70!black}{Arc-C}} &
   \parbox[t]{\columndistance}{\centering \textcolor{blue!70!black}{Wino}} &
   \parbox[t]{\columndistance}{\centering \textcolor{blue!70!black}{OpenQA}} &
   \parbox[t]{\columndistance}{\centering \textbf{\textcolor{teal}{Avg.}}} \\
\midrule
\multirow{7}{*}{\parbox[t]{1.5cm}{\centering LLaMA\\3.1 8B}} & Dense & -- & 5.84 & 63.57 & 80.09 & 81.44 & 51.37 & 73.48 & 33.40 & 63.89 \\
 & Wanda & -- & {21.65} & 31.98 & 69.53 & 61.11 & 27.30 & 61.09 & 21.40 & 45.40 \\
 & \cellcolor{shade}Wanda & \cellcolor{shade}\ours & \cellcolor{shade}{17.56} & \cellcolor{shade}{\textbf{33.96}} & \cellcolor{shade}{\textbf{71.60}} & \cellcolor{shade}{\textbf{63.76}} & \cellcolor{shade}{\textbf{29.35}} & \cellcolor{shade}{\textbf{66.06}} & \cellcolor{shade}{\textbf{22.60}} & \cellcolor{shade}{\textbf{47.89}} \\
 & SparseGPT & SparseGPT & 15.44 & \textbf{35.32} & 71.55 & 62.88 & 31.66 & \textbf{68.19} & 24.20 & \textbf{48.96} \\
 & \cellcolor{shade}SparseGPT & \cellcolor{shade}\ours & \cellcolor{shade}{{15.64}} & \cellcolor{shade}{32.44} & \cellcolor{shade}{\textbf{71.87}} & \cellcolor{shade}{\textbf{63.97}} & \cellcolor{shade}{\textbf{33.11}} & \cellcolor{shade}{67.56} & \cellcolor{shade}\textbf{24.60} & \cellcolor{shade}{48.93} \\
 & Thanos & Thanos & 15.91 & \textbf{35.22} & \textbf{72.09} & \textbf{65.28} & \textbf{33.19} & 67.40 & \textbf{23.40} & \textbf{49.43} \\
 & \cellcolor{shade}Thanos & \cellcolor{shade}\ours & \cellcolor{shade}{{16.09}} & \cellcolor{shade}{34.48} & \cellcolor{shade}{72.03} & \cellcolor{shade}{64.69} & \cellcolor{shade}{33.02} & \cellcolor{shade}\textbf{68.51} & \cellcolor{shade}{22.80} & \cellcolor{shade}{49.25} \\
\doublemidrule
\multirow{7}{*}{\parbox[t]{1.5cm}{\centering LLaMA\\3.2 1B}} & Dense & -- & 9.75 & 36.92 & 74.27 & 65.53 & 31.31 & 60.30 & 26.20 & 49.09 \\
 & Wanda & -- & {71.53} & 22.95 & 59.68 & 39.48 & 18.77 & 50.43 & 12.20 & 33.92 \\
 & \cellcolor{shade}Wanda & \cellcolor{shade}\ours & \cellcolor{shade}{41.50} & \cellcolor{shade}\textbf{23.52} & \cellcolor{shade}{\textbf{62.62}} & \cellcolor{shade}\textbf{44.53} & \cellcolor{shade}{\textbf{20.65}} & \cellcolor{shade}{\textbf{52.57}} & \cellcolor{shade}{\textbf{14.80}} & \cellcolor{shade}{\textbf{36.45}} \\
 & SparseGPT & SparseGPT & {48.00} & \textbf{23.02} & 62.08 & 43.48 & \textbf{21.76} & 52.09 & 17.40 & 36.64 \\
 & \cellcolor{shade}SparseGPT & \cellcolor{shade}\ours & \cellcolor{shade}{38.05} & \cellcolor{shade}{22.95} & \cellcolor{shade}\textbf{63.38} & \cellcolor{shade}{\textbf{43.52}} & \cellcolor{shade}{20.48} & \cellcolor{shade}{\textbf{53.28}} & \cellcolor{shade}\textbf{19.60} & \cellcolor{shade}\textbf{37.20} \\
 & Thanos & Thanos & {46.78} & \textbf{23.25} & 62.57 & 44.49 & 21.59 & 53.20 & 16.60 & 36.95 \\
 & \cellcolor{shade}Thanos & \cellcolor{shade}\ours & \cellcolor{shade}{40.54} & \cellcolor{shade}{23.02} & \cellcolor{shade}{\textbf{62.95}} & \cellcolor{shade}\textbf{44.53} & \cellcolor{shade}{\textbf{21.67}} & \cellcolor{shade}\textbf{53.91} & \cellcolor{shade}{\textbf{17.40}} & \cellcolor{shade}\textbf{37.25} \\
\doublemidrule
\multirow{7}{*}{\parbox[t]{1.5cm}{\centering LLaMA\\3.2 3B}} & Dense & -- & 7.81 & 54.13 & 76.55 & 74.28 & 42.75 & 69.38 & 30.60 & 57.95 \\
 & Wanda & -- & {31.13} & 25.53 & 65.23 & 47.90 & 22.70 & 55.25 & 16.00 & 38.77 \\
 & \cellcolor{shade}Wanda & \cellcolor{shade}\ours & \cellcolor{shade}{23.56} & \cellcolor{shade}{\textbf{31.20}} & \cellcolor{shade}{\textbf{67.41}} & \cellcolor{shade}{\textbf{53.96}} & \cellcolor{shade}{\textbf{24.57}} & \cellcolor{shade}{\textbf{59.51}} & \cellcolor{shade}{\textbf{19.80}} & \cellcolor{shade}{\textbf{42.74}} \\
 & SparseGPT & SparseGPT & 22.00 & \textbf{31.27} & \textbf{69.37} & 53.66 & \textbf{26.02} & 61.33 & \textbf{21.00} & \textbf{43.78} \\
 & \cellcolor{shade}SparseGPT & \cellcolor{shade}\ours & \cellcolor{shade}{{22.67}} & \cellcolor{shade}{29.58} & \cellcolor{shade}{68.77} & \cellcolor{shade}{\textbf{54.80}} & \cellcolor{shade}{24.74} & \cellcolor{shade}\textbf{62.35} & \cellcolor{shade}{20.60} & \cellcolor{shade}{43.47} \\
 & Thanos & Thanos & {22.48} & 29.23 & 67.63 & 55.01 & \textbf{26.02} & 57.85 & 19.20 & 42.49 \\
 & \cellcolor{shade}Thanos & \cellcolor{shade}\ours & \cellcolor{shade}{22.28} & \cellcolor{shade}\textbf{31.43} & \cellcolor{shade}{\textbf{67.90}} & \cellcolor{shade}\textbf{55.26} & \cellcolor{shade}{24.91} & \cellcolor{shade}{\textbf{59.67}} & \cellcolor{shade}{\textbf{20.60}} & \cellcolor{shade}{\textbf{43.30}} \\
\doublemidrule
\multirow{7}{*}{\parbox[t]{1.5cm}{\centering Gemma\\3 1B}} & Dense & -- & 14.17 & 24.95 & 74.81 & 71.93 & 35.41 & 58.72 & 28.80 & 49.10 \\
 & Wanda & -- & {90.48} & 23.04 & 62.19 & 49.75 & 18.60 & 50.99 & 15.20 & 36.63 \\
 & \cellcolor{shade}Wanda & \cellcolor{shade}\ours & \cellcolor{shade}64.79 & \cellcolor{shade}{\textbf{23.34}} & \cellcolor{shade}{\textbf{64.09}} & \cellcolor{shade}{\textbf{52.86}} & \cellcolor{shade}{\textbf{20.48}} & \cellcolor{shade}{\textbf{51.93}} & \cellcolor{shade}{\textbf{16.40}} & \cellcolor{shade}{\textbf{38.18}} \\
 & SparseGPT & SparseGPT & {60.91} & \textbf{24.58} & 65.34 & 51.98 & 21.93 & 51.14 & 16.60 & 38.60 \\
 & \cellcolor{shade}SparseGPT & \cellcolor{shade}\ours & \cellcolor{shade}{56.27} & \cellcolor{shade}{23.72} & \cellcolor{shade}\textbf{66.21} & \cellcolor{shade}{\textbf{52.44}} & \cellcolor{shade}\textbf{22.53} & \cellcolor{shade}{\textbf{52.96}} & \cellcolor{shade}{\textbf{17.60}} & \cellcolor{shade}{\textbf{39.24}} \\
 & Thanos & Thanos & {62.22} & \textbf{24.62} & 64.53 & 52.86 & 20.65 & 52.17 & 18.80 & 38.94 \\
 & \cellcolor{shade}Thanos & \cellcolor{shade}\ours & \cellcolor{shade}{56.78} & \cellcolor{shade}{24.44} & \cellcolor{shade}{\textbf{64.85}} & \cellcolor{shade}\textbf{55.18} & \cellcolor{shade}{\textbf{22.01}} & \cellcolor{shade}\textbf{54.85} & \cellcolor{shade}\textbf{19.80} & \cellcolor{shade}\textbf{40.19} \\
\doublemidrule
\multirow{7}{*}{\parbox[t]{1.5cm}{\centering Gemma\\2 2B}} & Dense & -- & 68.69 & 49.33 & 78.24 & 80.22 & 46.93 & 68.82 & 31.40 & 59.16 \\
 & Wanda & -- & {757.47} & 23.36 & 65.78 & 56.10 & 21.59 & 52.64 & 19.80 & 39.88 \\
 & \cellcolor{shade}Wanda & \cellcolor{shade}\ours & \cellcolor{shade}435.10 & \cellcolor{shade}{\textbf{24.37}} & \cellcolor{shade}{\textbf{66.59}} & \cellcolor{shade}\textbf{58.50} & \cellcolor{shade}{\textbf{21.93}} & \cellcolor{shade}{\textbf{57.38}} & \cellcolor{shade}{\textbf{20.00}} & \cellcolor{shade}{\textbf{41.46}} \\
 & SparseGPT & SparseGPT & {488.25} & 24.49 & 68.50 & 57.45 & 25.00 & \textbf{58.96} & \textbf{25.00} & 43.23 \\
 & \cellcolor{shade}SparseGPT & \cellcolor{shade}\ours & \cellcolor{shade}{451.46} & \cellcolor{shade}\textbf{25.89} & \cellcolor{shade}\textbf{68.88} & \cellcolor{shade}\textbf{58.50} & \cellcolor{shade}\textbf{26.28} & \cellcolor{shade}{58.01} & \cellcolor{shade}{24.20} & \cellcolor{shade}\textbf{43.63} \\
 & Thanos & Thanos & {523.61} & \textbf{23.69} & \textbf{68.23} & \textbf{58.12} & \textbf{23.89} & 58.33 & \textbf{21.20} & \textbf{42.24} \\
 & \cellcolor{shade}Thanos & \cellcolor{shade}\ours & \cellcolor{shade}{497.75} & \cellcolor{shade}{23.12} & \cellcolor{shade}{67.74} & \cellcolor{shade}{57.07} & \cellcolor{shade}{23.38} & \cellcolor{shade}\textbf{59.27} & \cellcolor{shade}{20.60} & \cellcolor{shade}{41.86} \\
\bottomrule
\end{tabular}
}
\caption{Model perplexity on WikiText2 and accuracy on zero-shot downstream tasks for 60\% unstructured sparsity. \ours improves the average accuracy in 11 of 15 (model, mask selector) pairs. Within each (Model, Mask Selection) block, the better of the two Weight Update rows is bolded.}
\label{tab:60_unstructured}
\end{table*}

\FloatBarrier
\section{Generalization to the Qwen-2.5 family}
\label{app:qwen}

To check that the results of \autoref{sec:results} carry over to a different model family, we repeat the experiments on Qwen-2.5 \citep{qwen2.5} at sizes from 0.5B to 14B parameters. \autoref{tab:qwen_50_unstructured}, \autoref{tab:qwen_60_unstructured}, and \autoref{tab:qwen_2_4_sparsity} report 50\% unstructured, 60\% unstructured, and 2:4 semi-structured results respectively.

The picture is consistent with LLaMA and Gemma at unstructured sparsity and mixed at 2:4. At 50\% unstructured sparsity, \ours improves 11 of the 15 (model size, mask selector) combinations (mean $+0.37$ points; the four regressions are between $-0.08$ and $-0.25$); for example, Qwen-2.5 3B pruned with Wanda goes from 54.02\% to 55.33\% average accuracy, and with SparseGPT from 54.70\% to 55.69\%. At 60\% sparsity the gains are larger, as expected (11 of 15 combinations improve, mean $+0.81$): on Qwen-2.5 3B, \ours lifts Wanda from 43.67\% to 47.86\% (a 4.19-point improvement) and Thanos from 48.45\% to 49.98\%. For 2:4 semi-structured pruning (MLP-only), 8 of the 9 pairs at 0.5B to 3B improve (Qwen-2.5 1.5B with Wanda from 46.01\% to 47.26\%, 3B from 49.48\% to 50.63\%), while the results regress for all three mask selectors on 7B (by 2.84 to 3.29 points) and are mixed on 14B (mean over all 15 pairs $-0.22$), suggesting that at this scale the interaction between structured masks and reconstruction is less uniform and would benefit from further study.

\begin{table}[tb]
\centering
\setlength{\tabcolsep}{2pt}
\resizebox{\textwidth}{!}{
\begin{tabular}{l l l c c c c c c c | c}
\toprule
\multirow{2}{*}{\parbox[t]{1.8cm}{\centering Model}} &
\multirow{2}{*}{\parbox[t]{1cm}{\centering Mask\\Selection}} &
\multirow{2}{*}{\parbox[t]{1cm}{\centering Weight\\Update}} &
\multirow{2}{*}{Perplexity} &
\multicolumn{7}{c}{Metrics (\%)} \\
\cmidrule{5-11}
 & & & &
  \parbox[t]{\columndistance}{\centering \textcolor{blue!70!black}{MMLU}} &
  \parbox[t]{\columndistance}{\centering \textcolor{blue!70!black}{PIQA}} &
  \parbox[t]{\columndistance}{\centering \textcolor{blue!70!black}{Arc-E}} &
  \parbox[t]{\columndistance}{\centering \textcolor{blue!70!black}{Arc-C}} &
  \parbox[t]{\columndistance}{\centering \textcolor{blue!70!black}{Wino}} &
  \parbox[t]{\columndistance}{\centering \textcolor{blue!70!black}{OpenQA}} &
  \parbox[t]{\columndistance}{\centering \textbf{\textcolor{teal}{Avg.}}} \\
\midrule

\multirow{7}{*}{\parbox[t]{1.8cm}{\centering Qwen 2.5\\0.5B}} & Dense & -- & 13.08 & 47.36 & 69.97 & 64.18 & 29.18 & 55.80 & 24.40 & 48.48 \\
 & Wanda & -- & 24.00 & \textbf{30.52} & 64.09 & 57.41 & 24.06 & 54.38 & 19.80 & 41.71 \\
 & \cellcolor{shade}{Wanda} & \cellcolor{shade}{\ours} & \cellcolor{shade}{\textbf{22.70}} & \cellcolor{shade}{26.14} & \cellcolor{shade}{\textbf{64.58}} & \cellcolor{shade}{\textbf{57.79}} & \cellcolor{shade}{\textbf{25.26}} & \cellcolor{shade}{\textbf{56.04}} & \cellcolor{shade}{\textbf{22.00}} & \cellcolor{shade}{\textbf{41.97}} \\
 & SparseGPT & SparseGPT & 20.33 & \textbf{29.38} & 64.74 & 56.52 & 24.15 & \textbf{56.20} & \textbf{20.60} & \textbf{41.93} \\
 & \cellcolor{shade}{SparseGPT} & \cellcolor{shade}{\ours} & \cellcolor{shade}{\textbf{19.54}} & \cellcolor{shade}{27.68} & \cellcolor{shade}{\textbf{65.13}} & \cellcolor{shade}{\textbf{56.99}} & \cellcolor{shade}{\textbf{24.66}} & \cellcolor{shade}{55.33} & \cellcolor{shade}{\textbf{20.60}} & \cellcolor{shade}{41.73} \\
 & Thanos & Thanos & 20.85 & 28.94 & \textbf{65.40} & 55.93 & \textbf{24.40} & \textbf{56.35} & 21.60 & 42.10 \\
 & \cellcolor{shade}{Thanos} & \cellcolor{shade}{\ours} & \cellcolor{shade}{\textbf{20.41}} & \cellcolor{shade}{\textbf{30.00}} & \cellcolor{shade}{64.69} & \cellcolor{shade}{\textbf{56.10}} & \cellcolor{shade}{\textbf{24.40}} & \cellcolor{shade}{55.41} & \cellcolor{shade}{\textbf{22.20}} & \cellcolor{shade}{\textbf{42.13}} \\
\doublemidrule

\multirow{7}{*}{\parbox[t]{1.8cm}{\centering Qwen 2.5\\1.5B}} & Dense & -- & 9.28 & 59.70 & 75.73 & 75.34 & 40.96 & 63.14 & 32.20 & 57.84 \\
 & Wanda & -- & 14.45 & 44.76 & 71.22 & \textbf{66.62} & 31.74 & 59.91 & \textbf{24.80} & 49.84 \\
 & \cellcolor{shade}{Wanda} & \cellcolor{shade}{\ours} & \cellcolor{shade}{\textbf{12.85}} & \cellcolor{shade}{\textbf{45.61}} & \cellcolor{shade}{\textbf{72.36}} & \cellcolor{shade}{\textbf{66.62}} & \cellcolor{shade}{\textbf{32.34}} & \cellcolor{shade}{\textbf{61.80}} & \cellcolor{shade}{24.60} & \cellcolor{shade}{\textbf{50.55}} \\
 & SparseGPT & SparseGPT & 13.09 & 46.80 & 71.65 & \textbf{66.75} & \textbf{33.62} & \textbf{62.27} & 25.60 & \textbf{51.12} \\
 & \cellcolor{shade}{SparseGPT} & \cellcolor{shade}{\ours} & \cellcolor{shade}{\textbf{12.76}} & \cellcolor{shade}{\textbf{46.96}} & \cellcolor{shade}{\textbf{71.82}} & \cellcolor{shade}{65.45} & \cellcolor{shade}{33.02} & \cellcolor{shade}{61.80} & \cellcolor{shade}{\textbf{26.20}} & \cellcolor{shade}{50.87} \\
 & Thanos & Thanos & 13.17 & \textbf{48.40} & 71.76 & 66.84 & \textbf{33.70} & \textbf{62.83} & \textbf{27.20} & \textbf{51.79} \\
 & \cellcolor{shade}{Thanos} & \cellcolor{shade}{\ours} & \cellcolor{shade}{\textbf{12.89}} & \cellcolor{shade}{48.21} & \cellcolor{shade}{\textbf{72.03}} & \cellcolor{shade}{\textbf{67.26}} & \cellcolor{shade}{33.53} & \cellcolor{shade}{62.04} & \cellcolor{shade}{26.20} & \cellcolor{shade}{51.55} \\
\doublemidrule

\multirow{7}{*}{\parbox[t]{1.8cm}{\centering Qwen 2.5\\3B}} & Dense & -- & 8.03 & 65.00 & 78.35 & 77.31 & 44.88 & 68.43 & 29.20 & 60.53 \\
 & Wanda & -- & 11.39 & 49.09 & 73.23 & 71.46 & \textbf{38.48} & 65.43 & 26.40 & 54.02 \\
 & \cellcolor{shade}{Wanda} & \cellcolor{shade}{\ours} & \cellcolor{shade}{\textbf{10.59}} & \cellcolor{shade}{\textbf{52.00}} & \cellcolor{shade}{\textbf{74.37}} & \cellcolor{shade}{\textbf{72.18}} & \cellcolor{shade}{38.05} & \cellcolor{shade}{\textbf{66.77}} & \cellcolor{shade}{\textbf{28.60}} & \cellcolor{shade}{\textbf{55.33}} \\
 & SparseGPT & SparseGPT & 10.74 & 52.49 & 74.65 & \textbf{71.34} & 36.86 & 64.64 & 28.20 & 54.70 \\
 & \cellcolor{shade}{SparseGPT} & \cellcolor{shade}{\ours} & \cellcolor{shade}{\textbf{10.57}} & \cellcolor{shade}{\textbf{53.92}} & \cellcolor{shade}{\textbf{75.35}} & \cellcolor{shade}{70.83} & \cellcolor{shade}{\textbf{38.31}} & \cellcolor{shade}{\textbf{66.14}} & \cellcolor{shade}{\textbf{29.60}} & \cellcolor{shade}{\textbf{55.69}} \\
 & Thanos & Thanos & 10.64 & \textbf{52.61} & \textbf{75.52} & \textbf{70.54} & 36.69 & 66.61 & \textbf{28.40} & \textbf{55.06} \\
 & \cellcolor{shade}{Thanos} & \cellcolor{shade}{\ours} & \cellcolor{shade}{\textbf{10.52}} & \cellcolor{shade}{52.11} & \cellcolor{shade}{75.46} & \cellcolor{shade}{70.12} & \cellcolor{shade}{\textbf{37.29}} & \cellcolor{shade}{\textbf{66.69}} & \cellcolor{shade}{28.20} & \cellcolor{shade}{54.98} \\
\doublemidrule

\multirow{7}{*}{\parbox[t]{1.8cm}{\centering Qwen 2.5\\7B}} & Dense & -- & 6.85 & 71.76 & 78.73 & 80.51 & 48.38 & 72.61 & 33.40 & 64.23 \\
 & Wanda & -- & 8.62 & 65.89 & 77.31 & 75.08 & 40.53 & 70.17 & \textbf{30.80} & 59.96 \\
 & \cellcolor{shade}{Wanda} & \cellcolor{shade}{\ours} & \cellcolor{shade}{\textbf{8.33}} & \cellcolor{shade}{\textbf{66.17}} & \cellcolor{shade}{\textbf{77.69}} & \cellcolor{shade}{\textbf{76.43}} & \cellcolor{shade}{\textbf{42.66}} & \cellcolor{shade}{\textbf{71.27}} & \cellcolor{shade}{30.60} & \cellcolor{shade}{\textbf{60.80}} \\
 & SparseGPT & SparseGPT & 8.42 & \textbf{66.09} & \textbf{78.07} & 75.34 & 42.75 & 71.11 & 31.00 & 60.73 \\
 & \cellcolor{shade}{SparseGPT} & \cellcolor{shade}{\ours} & \cellcolor{shade}{\textbf{8.36}} & \cellcolor{shade}{65.78} & \cellcolor{shade}{77.64} & \cellcolor{shade}{\textbf{75.63}} & \cellcolor{shade}{\textbf{42.92}} & \cellcolor{shade}{\textbf{71.51}} & \cellcolor{shade}{\textbf{31.60}} & \cellcolor{shade}{\textbf{60.85}} \\
 & Thanos & Thanos & 8.49 & 66.21 & \textbf{77.86} & 74.71 & 42.32 & 70.17 & 30.40 & 60.28 \\
 & \cellcolor{shade}{Thanos} & \cellcolor{shade}{\ours} & \cellcolor{shade}{\textbf{8.46}} & \cellcolor{shade}{\textbf{66.23}} & \cellcolor{shade}{77.58} & \cellcolor{shade}{\textbf{76.22}} & \cellcolor{shade}{\textbf{44.45}} & \cellcolor{shade}{\textbf{71.19}} & \cellcolor{shade}{\textbf{31.20}} & \cellcolor{shade}{\textbf{61.15}} \\
\doublemidrule

\multirow{7}{*}{\parbox[t]{1.8cm}{\centering Qwen 2.5\\14B}} & Dense & -- & 5.30 & 77.62 & 81.28 & 82.24 & 55.80 & 75.14 & 34.40 & 67.75 \\
 & Wanda & -- & 7.30 & \textbf{69.84} & 79.16 & 81.02 & 51.28 & 73.72 & \textbf{34.60} & 64.94 \\
 & \cellcolor{shade}{Wanda} & \cellcolor{shade}{\ours} & \cellcolor{shade}{\textbf{7.18}} & \cellcolor{shade}{69.29} & \cellcolor{shade}{\textbf{79.43}} & \cellcolor{shade}{\textbf{81.19}} & \cellcolor{shade}{\textbf{52.30}} & \cellcolor{shade}{\textbf{73.80}} & \cellcolor{shade}{33.80} & \cellcolor{shade}{\textbf{64.97}} \\
 & SparseGPT & SparseGPT & 7.24 & \textbf{69.83} & \textbf{79.60} & 80.98 & 51.02 & 72.93 & 32.80 & 64.53 \\
 & \cellcolor{shade}{SparseGPT} & \cellcolor{shade}{\ours} & \cellcolor{shade}{\textbf{7.14}} & \cellcolor{shade}{69.71} & \cellcolor{shade}{79.54} & \cellcolor{shade}{\textbf{81.19}} & \cellcolor{shade}{\textbf{51.79}} & \cellcolor{shade}{\textbf{73.80}} & \cellcolor{shade}{\textbf{33.60}} & \cellcolor{shade}{\textbf{64.94}} \\
 & Thanos & Thanos & 7.25 & \textbf{70.57} & \textbf{79.87} & 80.18 & 49.15 & 73.09 & 32.20 & 64.17 \\
 & \cellcolor{shade}{Thanos} & \cellcolor{shade}{\ours} & \cellcolor{shade}{\textbf{7.19}} & \cellcolor{shade}{70.16} & \cellcolor{shade}{79.60} & \cellcolor{shade}{\textbf{81.57}} & \cellcolor{shade}{\textbf{51.37}} & \cellcolor{shade}{\textbf{73.48}} & \cellcolor{shade}{\textbf{33.00}} & \cellcolor{shade}{\textbf{64.86}} \\
\bottomrule
\end{tabular}
}
\caption{Qwen-2.5 family perplexity on WikiText2 and accuracy on zero-shot downstream tasks for 50\% unstructured sparsity. \ours improves the average accuracy in 11 of 15 (model size, mask selector) pairs. Within each (Model, Mask Selection) block, the better of the two Weight Update rows is bolded.}
\label{tab:qwen_50_unstructured}
\end{table}

\begin{table}[tb]
\centering
\setlength{\tabcolsep}{2pt}
\resizebox{\textwidth}{!}{
\begin{tabular}{l l l c c c c c c c | c}
\toprule
\multirow{2}{*}{\parbox[t]{1.8cm}{\centering Model}} &
\multirow{2}{*}{\parbox[t]{1cm}{\centering Mask\\Selection}} &
\multirow{2}{*}{\parbox[t]{1cm}{\centering Weight\\Update}} &
\multirow{2}{*}{Perplexity} &
\multicolumn{7}{c}{Metrics (\%)} \\
\cmidrule{5-11}
 & & & &
  \parbox[t]{\columndistance}{\centering \textcolor{blue!70!black}{MMLU}} &
  \parbox[t]{\columndistance}{\centering \textcolor{blue!70!black}{PIQA}} &
  \parbox[t]{\columndistance}{\centering \textcolor{blue!70!black}{Arc-E}} &
  \parbox[t]{\columndistance}{\centering \textcolor{blue!70!black}{Arc-C}} &
  \parbox[t]{\columndistance}{\centering \textcolor{blue!70!black}{Wino}} &
  \parbox[t]{\columndistance}{\centering \textcolor{blue!70!black}{OpenQA}} &
  \parbox[t]{\columndistance}{\centering \textbf{\textcolor{teal}{Average}}} \\
\midrule

\multirow{7}{*}{\parbox[t]{1.8cm}{\centering Qwen 2.5\\0.5B}} & Dense & -- & 13.08 & 47.36 & 69.97 & 64.18 & 29.18 & 55.80 & 24.40 & 48.48 \\
 & Wanda & -- & 83.42 & 23.02 & 59.96 & 43.81 & 18.09 & 50.28 & 12.80 & 34.66 \\
 & \cellcolor{shade}{Wanda} & \cellcolor{shade}{\ours} & \cellcolor{shade}{\textbf{51.97}} & \cellcolor{shade}{\textbf{23.16}} & \cellcolor{shade}{\textbf{60.72}} & \cellcolor{shade}{\textbf{46.25}} & \cellcolor{shade}{\textbf{20.14}} & \cellcolor{shade}{\textbf{51.78}} & \cellcolor{shade}{\textbf{16.40}} & \cellcolor{shade}{\textbf{36.41}} \\
 & SparseGPT & SparseGPT & 40.56 & 22.90 & 61.59 & 48.40 & 21.25 & 52.80 & 16.80 & 37.29 \\
 & \cellcolor{shade}{SparseGPT} & \cellcolor{shade}{\ours} & \cellcolor{shade}{\textbf{36.77}} & \cellcolor{shade}{\textbf{23.06}} & \cellcolor{shade}{\textbf{62.13}} & \cellcolor{shade}{\textbf{48.74}} & \cellcolor{shade}{\textbf{21.33}} & \cellcolor{shade}{\textbf{53.99}} & \cellcolor{shade}{\textbf{17.40}} & \cellcolor{shade}{\textbf{37.77}} \\
 & Thanos & Thanos & 44.29 & \textbf{23.78} & \textbf{62.02} & \textbf{48.65} & 21.33 & 52.25 & 17.80 & 37.64 \\
 & \cellcolor{shade}{Thanos} & \cellcolor{shade}{\ours} & \cellcolor{shade}{\textbf{41.92}} & \cellcolor{shade}{23.59} & \cellcolor{shade}{61.86} & \cellcolor{shade}{46.80} & \cellcolor{shade}{\textbf{22.35}} & \cellcolor{shade}{\textbf{53.75}} & \cellcolor{shade}{\textbf{19.60}} & \cellcolor{shade}{\textbf{37.99}} \\
\doublemidrule

\multirow{7}{*}{\parbox[t]{1.8cm}{\centering Qwen 2.5\\1.5B}} & Dense & -- & 9.28 & 59.70 & 75.73 & 75.34 & 40.96 & 63.14 & 32.20 & 57.84 \\
 & Wanda & -- & 58.38 & 27.25 & 65.18 & 54.50 & 24.74 & 53.04 & 17.20 & 40.32 \\
 & \cellcolor{shade}{Wanda} & \cellcolor{shade}{\ours} & \cellcolor{shade}{\textbf{23.81}} & \cellcolor{shade}{\textbf{30.99}} & \cellcolor{shade}{\textbf{66.87}} & \cellcolor{shade}{\textbf{56.44}} & \cellcolor{shade}{\textbf{24.91}} & \cellcolor{shade}{\textbf{56.83}} & \cellcolor{shade}{\textbf{18.40}} & \cellcolor{shade}{\textbf{42.41}} \\
 & SparseGPT & SparseGPT & 21.92 & \textbf{33.56} & 67.36 & \textbf{58.08} & \textbf{27.47} & 57.14 & 21.60 & \textbf{44.20} \\
 & \cellcolor{shade}{SparseGPT} & \cellcolor{shade}{\ours} & \cellcolor{shade}{\textbf{19.35}} & \cellcolor{shade}{31.44} & \cellcolor{shade}{\textbf{67.79}} & \cellcolor{shade}{56.27} & \cellcolor{shade}{27.22} & \cellcolor{shade}{\textbf{59.27}} & \cellcolor{shade}{\textbf{22.40}} & \cellcolor{shade}{44.07} \\
 & Thanos & Thanos & 27.07 & 33.66 & \textbf{67.63} & 57.49 & \textbf{27.73} & 56.67 & 20.60 & 43.96 \\
 & \cellcolor{shade}{Thanos} & \cellcolor{shade}{\ours} & \cellcolor{shade}{\textbf{23.64}} & \cellcolor{shade}{\textbf{35.97}} & \cellcolor{shade}{67.14} & \cellcolor{shade}{\textbf{57.66}} & \cellcolor{shade}{26.19} & \cellcolor{shade}{\textbf{58.09}} & \cellcolor{shade}{\textbf{20.80}} & \cellcolor{shade}{\textbf{44.31}} \\
\doublemidrule

\multirow{7}{*}{\parbox[t]{1.8cm}{\centering Qwen 2.5\\3B}} & Dense & -- & 8.03 & 65.00 & 78.35 & 77.31 & 44.88 & 68.43 & 29.20 & 60.53 \\
 & Wanda & -- & 22.06 & 28.07 & 67.14 & 60.86 & 27.39 & 58.17 & 20.40 & 43.67 \\
 & \cellcolor{shade}{Wanda} & \cellcolor{shade}{\ours} & \cellcolor{shade}{\textbf{15.67}} & \cellcolor{shade}{\textbf{37.22}} & \cellcolor{shade}{\textbf{70.24}} & \cellcolor{shade}{\textbf{63.55}} & \cellcolor{shade}{\textbf{30.89}} & \cellcolor{shade}{\textbf{61.64}} & \cellcolor{shade}{\textbf{23.60}} & \cellcolor{shade}{\textbf{47.86}} \\
 & SparseGPT & SparseGPT & 14.82 & \textbf{43.16} & 71.60 & 64.35 & 32.59 & 63.30 & 23.20 & 49.70 \\
 & \cellcolor{shade}{SparseGPT} & \cellcolor{shade}{\ours} & \cellcolor{shade}{\textbf{14.50}} & \cellcolor{shade}{40.25} & \cellcolor{shade}{\textbf{72.20}} & \cellcolor{shade}{\textbf{64.90}} & \cellcolor{shade}{\textbf{33.62}} & \cellcolor{shade}{\textbf{63.69}} & \cellcolor{shade}{\textbf{24.00}} & \cellcolor{shade}{\textbf{49.78}} \\
 & Thanos & Thanos & 14.90 & 40.76 & \textbf{71.38} & 63.26 & 30.63 & 61.25 & 23.40 & 48.45 \\
 & \cellcolor{shade}{Thanos} & \cellcolor{shade}{\ours} & \cellcolor{shade}{\textbf{14.42}} & \cellcolor{shade}{\textbf{42.58}} & \cellcolor{shade}{71.27} & \cellcolor{shade}{\textbf{64.73}} & \cellcolor{shade}{\textbf{32.68}} & \cellcolor{shade}{\textbf{63.61}} & \cellcolor{shade}{\textbf{25.00}} & \cellcolor{shade}{\textbf{49.98}} \\
\doublemidrule

\multirow{7}{*}{\parbox[t]{1.8cm}{\centering Qwen 2.5\\7B}} & Dense & -- & 6.85 & 71.76 & 78.73 & 80.51 & 48.38 & 72.61 & 33.40 & 64.23 \\
 & Wanda & -- & 14.09 & 54.58 & 72.03 & 71.68 & 37.03 & 66.46 & 25.40 & 54.53 \\
 & \cellcolor{shade}{Wanda} & \cellcolor{shade}{\ours} & \cellcolor{shade}{\textbf{11.15}} & \cellcolor{shade}{\textbf{55.49}} & \cellcolor{shade}{\textbf{73.99}} & \cellcolor{shade}{\textbf{73.86}} & \cellcolor{shade}{\textbf{37.88}} & \cellcolor{shade}{\textbf{67.96}} & \cellcolor{shade}{\textbf{26.20}} & \cellcolor{shade}{\textbf{55.90}} \\
 & SparseGPT & SparseGPT & 10.86 & \textbf{56.63} & 74.92 & 73.36 & 40.61 & \textbf{67.25} & 25.80 & 56.43 \\
 & \cellcolor{shade}{SparseGPT} & \cellcolor{shade}{\ours} & \cellcolor{shade}{\textbf{10.53}} & \cellcolor{shade}{55.55} & \cellcolor{shade}{\textbf{75.46}} & \cellcolor{shade}{\textbf{73.78}} & \cellcolor{shade}{\textbf{40.70}} & \cellcolor{shade}{66.93} & \cellcolor{shade}{\textbf{26.60}} & \cellcolor{shade}{\textbf{56.50}} \\
 & Thanos & Thanos & 11.07 & \textbf{59.54} & 74.70 & \textbf{73.44} & \textbf{40.44} & 69.22 & 26.40 & \textbf{57.29} \\
 & \cellcolor{shade}{Thanos} & \cellcolor{shade}{\ours} & \cellcolor{shade}{\textbf{10.74}} & \cellcolor{shade}{58.90} & \cellcolor{shade}{\textbf{75.35}} & \cellcolor{shade}{72.69} & \cellcolor{shade}{40.10} & \cellcolor{shade}{\textbf{69.46}} & \cellcolor{shade}{\textbf{27.00}} & \cellcolor{shade}{57.25} \\
\doublemidrule

\multirow{7}{*}{\parbox[t]{1.8cm}{\centering Qwen 2.5\\14B}} & Dense & -- & 5.30 & 77.62 & 81.28 & 82.24 & 55.80 & 75.14 & 34.40 & 67.75 \\
 & Wanda & -- & 11.16 & 61.38 & 75.41 & 74.12 & \textbf{42.15} & 71.51 & 29.20 & 58.96 \\
 & \cellcolor{shade}{Wanda} & \cellcolor{shade}{\ours} & \cellcolor{shade}{\textbf{9.69}} & \cellcolor{shade}{\textbf{61.74}} & \cellcolor{shade}{\textbf{75.57}} & \cellcolor{shade}{\textbf{75.34}} & \cellcolor{shade}{41.98} & \cellcolor{shade}{\textbf{73.09}} & \cellcolor{shade}{\textbf{29.40}} & \cellcolor{shade}{\textbf{59.52}} \\
 & SparseGPT & SparseGPT & 9.22 & \textbf{62.83} & 76.66 & 76.18 & 44.45 & \textbf{72.14} & \textbf{29.60} & \textbf{60.31} \\
 & \cellcolor{shade}{SparseGPT} & \cellcolor{shade}{\ours} & \cellcolor{shade}{\textbf{8.97}} & \cellcolor{shade}{62.22} & \cellcolor{shade}{\textbf{76.93}} & \cellcolor{shade}{\textbf{76.47}} & \cellcolor{shade}{\textbf{44.54}} & \cellcolor{shade}{71.67} & \cellcolor{shade}{29.00} & \cellcolor{shade}{60.14} \\
 & Thanos & Thanos & 9.14 & \textbf{63.03} & \textbf{77.20} & 76.05 & 43.77 & 71.98 & 29.80 & \textbf{60.31} \\
 & \cellcolor{shade}{Thanos} & \cellcolor{shade}{\ours} & \cellcolor{shade}{\textbf{8.99}} & \cellcolor{shade}{60.30} & \cellcolor{shade}{76.39} & \cellcolor{shade}{\textbf{76.30}} & \cellcolor{shade}{\textbf{43.77}} & \cellcolor{shade}{\textbf{72.14}} & \cellcolor{shade}{\textbf{30.60}} & \cellcolor{shade}{59.92} \\
\bottomrule
\end{tabular}
}
\caption{Qwen-2.5 perplexity on WikiText2 and accuracy on zero-shot downstream tasks for 60\% unstructured sparsity. \ours improves the average accuracy in 11 of 15 (model size, mask selector) pairs. Within each (Model, Mask Selection) block, the better of the two Weight Update rows is bolded.}
\label{tab:qwen_60_unstructured}
\end{table}

\begin{table}[tb]
\centering
\setlength{\tabcolsep}{2pt}
\resizebox{\textwidth}{!}{
\begin{tabular}{l l l c c c c c c c | c}
\toprule
\multirow{2}{*}{\parbox[t]{1.8cm}{\centering Model}} &
\multirow{2}{*}{\parbox[t]{1cm}{\centering Mask\\Selection}} &
\multirow{2}{*}{\parbox[t]{1cm}{\centering Weight\\Update}} &
\multirow{2}{*}{Perplexity} &
\multicolumn{7}{c}{Metrics (\%)} \\
\cmidrule{5-11}
 & & & &
  \parbox[t]{\columndistance}{\centering \textcolor{blue!70!black}{MMLU}} &
  \parbox[t]{\columndistance}{\centering \textcolor{blue!70!black}{PIQA}} &
  \parbox[t]{\columndistance}{\centering \textcolor{blue!70!black}{Arc-E}} &
  \parbox[t]{\columndistance}{\centering \textcolor{blue!70!black}{Arc-C}} &
  \parbox[t]{\columndistance}{\centering \textcolor{blue!70!black}{Wino}} &
  \parbox[t]{\columndistance}{\centering \textcolor{blue!70!black}{OpenQA}} &
  \parbox[t]{\columndistance}{\centering \textbf{\textcolor{teal}{Avg.}}} \\
\midrule

\multirow{7}{*}{\parbox[t]{1.8cm}{\centering Qwen 2.5\\0.5B}} & Dense & -- & 13.08 & 47.36 & 69.97 & 64.18 & 29.18 & 55.80 & 24.40 & 48.48 \\
 & Wanda & -- & 41.30 & \textbf{27.88} & 61.75 & 48.99 & \textbf{23.38} & 52.72 & 14.00 & 38.12 \\
 & \cellcolor{shade}{Wanda} & \cellcolor{shade}{\ours} & \cellcolor{shade}{\textbf{27.61}} & \cellcolor{shade}{25.57} & \cellcolor{shade}{\textbf{63.76}} & \cellcolor{shade}{\textbf{51.18}} & \cellcolor{shade}{22.18} & \cellcolor{shade}{\textbf{53.28}} & \cellcolor{shade}{\textbf{15.80}} & \cellcolor{shade}{\textbf{38.63}} \\
 & SparseGPT & SparseGPT & 27.15 & \textbf{24.83} & 62.79 & 49.41 & 22.35 & 52.33 & \textbf{17.20} & 38.15 \\
 & \cellcolor{shade}{SparseGPT} & \cellcolor{shade}{\ours} & \cellcolor{shade}{\textbf{25.77}} & \cellcolor{shade}{23.38} & \cellcolor{shade}{\textbf{62.95}} & \cellcolor{shade}{\textbf{51.30}} & \cellcolor{shade}{\textbf{22.95}} & \cellcolor{shade}{\textbf{54.70}} & \cellcolor{shade}{17.00} & \cellcolor{shade}{\textbf{38.71}} \\
 & Thanos & Thanos & 27.58 & \textbf{24.31} & 62.68 & 49.92 & 21.42 & 51.78 & \textbf{16.60} & 37.78 \\
 & \cellcolor{shade}{Thanos} & \cellcolor{shade}{\ours} & \cellcolor{shade}{\textbf{26.26}} & \cellcolor{shade}{23.60} & \cellcolor{shade}{\textbf{62.79}} & \cellcolor{shade}{\textbf{51.22}} & \cellcolor{shade}{\textbf{22.18}} & \cellcolor{shade}{\textbf{54.30}} & \cellcolor{shade}{\textbf{16.60}} & \cellcolor{shade}{\textbf{38.45}} \\
\doublemidrule

\multirow{7}{*}{\parbox[t]{1.8cm}{\centering Qwen 2.5\\1.5B}} & Dense & -- & 9.28 & 59.70 & 75.73 & 75.34 & 40.96 & 63.14 & 32.20 & 57.84 \\
 & Wanda & -- & 21.92 & 39.95 & 67.25 & 61.11 & 28.84 & 58.09 & 20.80 & 46.01 \\
 & \cellcolor{shade}{Wanda} & \cellcolor{shade}{\ours} & \cellcolor{shade}{\textbf{17.14}} & \cellcolor{shade}{\textbf{39.96}} & \cellcolor{shade}{\textbf{69.26}} & \cellcolor{shade}{\textbf{62.33}} & \cellcolor{shade}{\textbf{30.29}} & \cellcolor{shade}{\textbf{58.72}} & \cellcolor{shade}{\textbf{23.00}} & \cellcolor{shade}{\textbf{47.26}} \\
 & SparseGPT & SparseGPT & 17.24 & \textbf{41.05} & 69.64 & \textbf{63.09} & \textbf{30.63} & \textbf{61.17} & \textbf{23.00} & \textbf{48.10} \\
 & \cellcolor{shade}{SparseGPT} & \cellcolor{shade}{\ours} & \cellcolor{shade}{\textbf{16.52}} & \cellcolor{shade}{35.64} & \cellcolor{shade}{\textbf{69.91}} & \cellcolor{shade}{62.75} & \cellcolor{shade}{29.35} & \cellcolor{shade}{60.93} & \cellcolor{shade}{\textbf{23.00}} & \cellcolor{shade}{46.93} \\
 & Thanos & Thanos & 17.56 & \textbf{42.83} & 68.55 & 61.53 & 29.10 & \textbf{58.96} & 21.40 & 47.06 \\
 & \cellcolor{shade}{Thanos} & \cellcolor{shade}{\ours} & \cellcolor{shade}{\textbf{16.83}} & \cellcolor{shade}{40.31} & \cellcolor{shade}{\textbf{69.75}} & \cellcolor{shade}{\textbf{62.67}} & \cellcolor{shade}{\textbf{30.38}} & \cellcolor{shade}{58.56} & \cellcolor{shade}{\textbf{25.00}} & \cellcolor{shade}{\textbf{47.78}} \\
\doublemidrule

\multirow{7}{*}{\parbox[t]{1.8cm}{\centering Qwen 2.5\\3B}} & Dense & -- & 8.03 & 65.00 & 78.35 & 77.31 & 44.88 & 68.43 & 29.20 & 60.53 \\
 & Wanda & -- & 17.14 & \textbf{46.68} & 70.08 & 64.77 & \textbf{31.66} & 61.48 & 22.20 & 49.48 \\
 & \cellcolor{shade}{Wanda} & \cellcolor{shade}{\ours} & \cellcolor{shade}{\textbf{14.08}} & \cellcolor{shade}{46.55} & \cellcolor{shade}{\textbf{71.44}} & \cellcolor{shade}{\textbf{64.90}} & \cellcolor{shade}{31.31} & \cellcolor{shade}{\textbf{64.17}} & \cellcolor{shade}{\textbf{25.40}} & \cellcolor{shade}{\textbf{50.63}} \\
 & SparseGPT & SparseGPT & 14.06 & \textbf{43.79} & \textbf{72.03} & 66.16 & 31.31 & 64.96 & 25.20 & 50.58 \\
 & \cellcolor{shade}{SparseGPT} & \cellcolor{shade}{\ours} & \cellcolor{shade}{\textbf{13.57}} & \cellcolor{shade}{43.36} & \cellcolor{shade}{71.44} & \cellcolor{shade}{\textbf{66.62}} & \cellcolor{shade}{\textbf{31.91}} & \cellcolor{shade}{\textbf{65.27}} & \cellcolor{shade}{\textbf{27.00}} & \cellcolor{shade}{\textbf{50.93}} \\
 & Thanos & Thanos & 14.35 & 41.41 & \textbf{71.49} & 61.70 & 29.27 & 64.01 & 25.20 & 48.85 \\
 & \cellcolor{shade}{Thanos} & \cellcolor{shade}{\ours} & \cellcolor{shade}{\textbf{13.75}} & \cellcolor{shade}{\textbf{43.56}} & \cellcolor{shade}{70.73} & \cellcolor{shade}{\textbf{64.06}} & \cellcolor{shade}{\textbf{30.97}} & \cellcolor{shade}{\textbf{64.09}} & \cellcolor{shade}{\textbf{25.40}} & \cellcolor{shade}{\textbf{49.80}} \\
\doublemidrule

\multirow{7}{*}{\parbox[t]{1.8cm}{\centering Qwen 2.5\\7B}} & Dense & -- & 6.85 & 71.76 & 78.73 & 80.51 & 48.38 & 72.61 & 33.40 & 64.23 \\
 & Wanda & -- & \textbf{11.47} & \textbf{61.03} & \textbf{74.48} & \textbf{75.34} & \textbf{42.15} & \textbf{68.82} & \textbf{27.80} & \textbf{58.27} \\
 & \cellcolor{shade}{Wanda} & \cellcolor{shade}{\ours} & \cellcolor{shade}{11.80} & \cellcolor{shade}{53.90} & \cellcolor{shade}{74.10} & \cellcolor{shade}{71.97} & \cellcolor{shade}{38.05} & \cellcolor{shade}{67.96} & \cellcolor{shade}{26.40} & \cellcolor{shade}{55.40} \\
 & SparseGPT & SparseGPT & \textbf{10.21} & \textbf{60.30} & \textbf{75.57} & \textbf{75.59} & \textbf{41.38} & \textbf{71.51} & \textbf{28.20} & \textbf{58.76} \\
 & \cellcolor{shade}{SparseGPT} & \cellcolor{shade}{\ours} & \cellcolor{shade}{10.92} & \cellcolor{shade}{53.90} & \cellcolor{shade}{74.32} & \cellcolor{shade}{72.43} & \cellcolor{shade}{37.46} & \cellcolor{shade}{69.61} & \cellcolor{shade}{27.80} & \cellcolor{shade}{55.92} \\
 & Thanos & Thanos & \textbf{10.45} & \textbf{60.12} & \textbf{74.54} & \textbf{75.08} & \textbf{41.13} & \textbf{69.93} & \textbf{28.80} & \textbf{58.27} \\
 & \cellcolor{shade}{Thanos} & \cellcolor{shade}{\ours} & \cellcolor{shade}{11.13} & \cellcolor{shade}{54.90} & \cellcolor{shade}{73.94} & \cellcolor{shade}{70.83} & \cellcolor{shade}{35.15} & \cellcolor{shade}{69.06} & \cellcolor{shade}{26.00} & \cellcolor{shade}{54.98} \\
\doublemidrule

\multirow{7}{*}{\parbox[t]{1.8cm}{\centering Qwen 2.5\\14B}} & Dense & -- & 5.30 & 77.62 & 81.28 & 82.24 & 55.80 & 75.14 & 34.40 & 67.75 \\
 & Wanda & -- & 9.70 & 65.82 & 76.99 & 76.89 & 45.39 & 73.56 & 31.80 & 61.74 \\
 & \cellcolor{shade}{Wanda} & \cellcolor{shade}{\ours} & \cellcolor{shade}{\textbf{8.90}} & \cellcolor{shade}{\textbf{67.03}} & \cellcolor{shade}{\textbf{77.31}} & \cellcolor{shade}{\textbf{77.82}} & \cellcolor{shade}{\textbf{46.76}} & \cellcolor{shade}{\textbf{74.27}} & \cellcolor{shade}{\textbf{32.60}} & \cellcolor{shade}{\textbf{62.63}} \\
 & SparseGPT & SparseGPT & 9.02 & \textbf{67.45} & 77.58 & 77.61 & \textbf{44.62} & 73.88 & \textbf{32.60} & \textbf{62.29} \\
 & \cellcolor{shade}{SparseGPT} & \cellcolor{shade}{\ours} & \cellcolor{shade}{\textbf{8.82}} & \cellcolor{shade}{67.33} & \cellcolor{shade}{\textbf{77.58}} & \cellcolor{shade}{\textbf{77.82}} & \cellcolor{shade}{44.28} & \cellcolor{shade}{\textbf{73.88}} & \cellcolor{shade}{32.00} & \cellcolor{shade}{62.15} \\
 & Thanos & Thanos & 9.06 & \textbf{66.31} & 77.64 & \textbf{77.90} & \textbf{46.25} & 72.77 & 31.20 & \textbf{62.01} \\
 & \cellcolor{shade}{Thanos} & \cellcolor{shade}{\ours} & \cellcolor{shade}{\textbf{8.92}} & \cellcolor{shade}{66.07} & \cellcolor{shade}{\textbf{77.97}} & \cellcolor{shade}{77.44} & \cellcolor{shade}{45.65} & \cellcolor{shade}{\textbf{72.93}} & \cellcolor{shade}{\textbf{31.80}} & \cellcolor{shade}{61.98} \\
\bottomrule
\end{tabular}
}
\caption{Qwen-2.5 perplexity on WikiText2 and accuracy on zero-shot downstream tasks for 2:4 sparsity. In this experiment, only the layers in the MLP part of the transformer are pruned, and the self-attention layers are dense, resulting in an end-to-end sparsity ratio of 38\% to 41\%. \ours improves the average accuracy in 9 of 15 (model size, mask selector) pairs. Within each (Model, Mask Selection) block, the better of the two Weight Update rows is bolded.}
\label{tab:qwen_2_4_sparsity}
\end{table}

\FloatBarrier
\section{Sensitivity to the calibration data}
\label{app:calibration}

\subsection{Variance over independent calibration draws}
\label{app:calibration_variance}

\ours introduces no stochasticity of its own: the fixed-mask problem is a convex QP solved to a global minimizer at the set tolerance, so re-running with different seeds reproduces the updates up to that tolerance and floating-point summation order. The only stochasticity in the pipeline is the calibration draw. \autoref{tab:calibration_variance} quantifies it directly with five independent draws of 128 C4 samples on LLaMA 3.2 1B at 50\% unstructured sparsity. The draw-to-draw standard deviation of the six-task average is at most 0.18 points, while \ours's gain over the Wanda baseline is $+1.39$ points, nearly eight times that standard deviation; the perplexity improvement (23.47 to 18.77) likewise dwarfs its standard deviations ($\leq 0.11$). The headline Wanda-mask gains are therefore far outside calibration-draw noise. Taking this standard deviation as representative, the smallest 50\% cells in \autoref{tab:50_unstructured} (e.g., the tie on LLaMA 3.1 8B with SparseGPT masks) are comparable to the noise scale, as \autoref{sec:results} concedes. All other results in this paper use matched calibration data; this experiment is the only one that varies the calibration samples.

\begin{table*}[tb]
\centering
\setlength{\tabcolsep}{2pt}
\resizebox{\textwidth}{!}{
\begin{tabular}{l l c c c c c c c | c}
\toprule
\multirow{2}{*}{\parbox[t]{1.6cm}{\centering Mask\\Selection}} &
\multirow{2}{*}{\parbox[t]{1.6cm}{\centering Weight\\Update}} &
\multirow{2}{*}{Perplexity} &
\multicolumn{7}{c}{Metrics (\%)} \\
\cmidrule{4-10}
 & & &
   \parbox[t]{\columndistance}{\centering \textcolor{blue!70!black}{MMLU}} &
   \parbox[t]{\columndistance}{\centering \textcolor{blue!70!black}{PIQA}} &
   \parbox[t]{\columndistance}{\centering \textcolor{blue!70!black}{Arc-E}} &
   \parbox[t]{\columndistance}{\centering \textcolor{blue!70!black}{Arc-C}} &
   \parbox[t]{\columndistance}{\centering \textcolor{blue!70!black}{Wino}} &
   \parbox[t]{\columndistance}{\centering \textcolor{blue!70!black}{OpenQA}} &
   \parbox[t]{\columndistance}{\centering \textbf{\textcolor{teal}{Avg.}}} \\
\midrule
Wanda & -- & 23.47 $\pm$ 0.11 & 26.38 $\pm$ 0.06 & 64.97 $\pm$ 0.17 & 51.96 $\pm$ 0.02 & 23.74 $\pm$ 0.06 & 54.66 $\pm$ 0.23 & 18.04 $\pm$ 0.08 & 39.96 $\pm$ 0.05 \\
\cellcolor{shade}Wanda & \cellcolor{shade}\ours & \cellcolor{shade}18.77 $\pm$ 0.09 & \cellcolor{shade}27.65 $\pm$ 0.13 & \cellcolor{shade}67.11 $\pm$ 0.27 & \cellcolor{shade}52.57 $\pm$ 0.05 & \cellcolor{shade}24.74 $\pm$ 0.10 & \cellcolor{shade}55.89 $\pm$ 0.13 & \cellcolor{shade}20.16 $\pm$ 0.08 & \cellcolor{shade}\textbf{41.35 $\pm$ 0.18} \\
\bottomrule
\end{tabular}
}
\caption{Mean $\pm$ standard deviation over five independent draws of 128 C4 calibration samples for LLaMA 3.2 1B at 50\% unstructured sparsity.}
\label{tab:calibration_variance}
\end{table*}

\subsection{Calibration dataset size sensitivity}
\label{app:calibraion_size}

Similar to previous work (SparseGPT, Wanda, Thanos), \ours leverages a set of calibration data from the C4 dataset to prune the models. \autoref{fig:calibration_size} shows the perplexity of LLaMA-3.2-1B on WikiText2 dataset when pruning the models with 32 to 256 calibration samples. \ours's perplexity stays below Wanda's at every size from 32 to 256 samples and changes little with the sample count, whereas Wanda and SparseGPT fluctuate non-monotonically, suggesting robustness to the calibration size.

\begin{figure}
    \centering
    \includegraphics{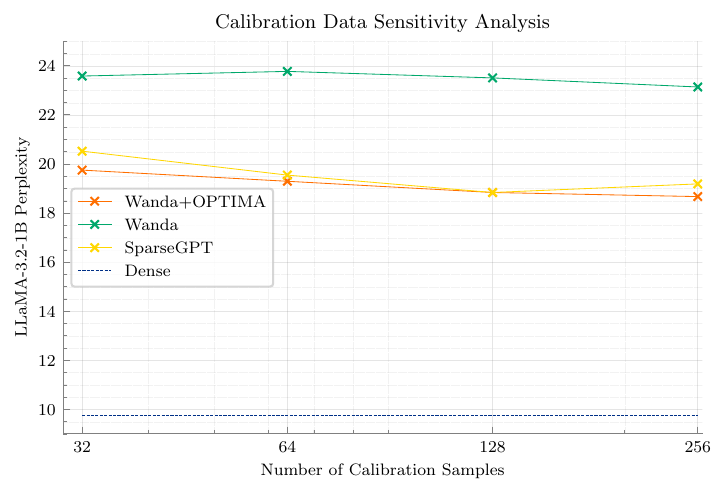}
    \caption{Sensitivity analysis for the number of calibration samples for different pruning methods.}
    \label{fig:calibration_size}
\end{figure}

\FloatBarrier
\section{Gradient-based alternatives to the QP solver}
\label{app:gradient_alternatives}

\subsection{ADAM as the column-wise solver}
\label{app:adam}

\autoref{sec:results} summarizes this experiment; the full per-model numbers are reported here.

While our constrained QP solver leverages theoretical guarantees for convergence and optimality, we also compare it against ADAM \citep{adam}, a popular first-order optimizer without such assurances for quadratic problems. We reformulate the weight update as a mean squared error (MSE) minimization problem and use ADAM for solving it. To keep the sparsity pattern fixed during this optimization, the pruned entries are re-zeroed after every step, so only the unpruned elements of the weight matrix are updated by ADAM and the pruned positions remain exactly zero throughout optimization.

Optimizers such as ADAM do not guarantee convergence, and are sensitive to their hyperparameters. For each layer, we do an exhaustive search with 4 different learning rates ranging from $10^{-2}$ to $10^{-5}$, each run for 20 steps with a linear learning rate scheduler decaying to zero, and choose the best configuration for the final weight update.

\begin{table}[tb]
\centering
\setlength{\tabcolsep}{2pt}
\resizebox{\textwidth}{!}{
\begin{tabular}{l l l c c c c c c c | c}
\toprule
\multirow{2}{*}{\parbox[t]{1.5cm}{\centering Model}} &
\multirow{2}{*}{\parbox[t]{1cm}{\centering Mask\\Selection}} &
\multirow{2}{*}{\parbox[t]{1cm}{\centering Weight\\Update}} &
\multirow{2}{*}{Perplexity} &
\multicolumn{7}{c}{Metrics (\%)} \\
\cmidrule{5-11}
 & & & &
   \parbox[t]{\columndistance}{\centering \textcolor{blue!70!black}{MMLU}} &
   \parbox[t]{\columndistance}{\centering \textcolor{blue!70!black}{PIQA}} &
   \parbox[t]{\columndistance}{\centering \textcolor{blue!70!black}{Arc-E}} &
   \parbox[t]{\columndistance}{\centering \textcolor{blue!70!black}{Arc-C}} &
   \parbox[t]{\columndistance}{\centering \textcolor{blue!70!black}{Wino}} &
   \parbox[t]{\columndistance}{\centering \textcolor{blue!70!black}{OpenQA}} &
   \parbox[t]{\columndistance}{\centering \textbf{\textcolor{teal}{Avg.}}} \\
\midrule
\multirow{7}{*}{\parbox[t]{1.5cm}{\centering Gemma\\3 1B}} & Dense & -- & 14.17 & 24.95 & 74.81 & 71.93 & 35.41 & 58.72 & 28.80 & 49.10 \\
 & Wanda & -- & 32.96 & 22.97 & 67.19 & 61.03 & 26.37 & 55.72 & 20.00 & 42.21 \\
 & Wanda & ADAM & 29.25 & 23.16 & 69.04 & 62.71 & 27.73 & 57.46 & 22.20 & 43.72 \\
 & \cellcolor{shade}Wanda & \cellcolor{shade}\ours & \cellcolor{shade}28.90 & \cellcolor{shade}23.96 & \cellcolor{shade}69.48 & \cellcolor{shade}\textbf{62.84} & \cellcolor{shade}\textbf{28.58} & \cellcolor{shade}56.83 & \cellcolor{shade}22.40 & \cellcolor{shade}\textbf{44.01} \\
 & SparseGPT & SparseGPT & 28.34 & 24.85 & 68.88 & 60.94 & 26.62 & 55.49 & 21.40 & 43.03 \\
 & SparseGPT & ADAM & 27.12 & 24.74 & 69.53 & 61.36 & 27.05 & 54.78 & 22.20 & 43.28 \\
 & \cellcolor{shade}SparseGPT & \cellcolor{shade}\ours & \cellcolor{shade}27.35 & \cellcolor{shade}\textbf{25.73} & \cellcolor{shade}\textbf{69.75} & \cellcolor{shade}60.90 & \cellcolor{shade}27.82 & \cellcolor{shade}56.35 & \cellcolor{shade}22.00 & \cellcolor{shade}43.76 \\
\doublemidrule
\multirow{7}{*}{\parbox[t]{1.5cm}{\centering OPT\\125M}} & Dense & -- & 27.67 & 22.85 & 62.84 & 43.56 & 19.45 & 49.88 & 16.40 & 35.83 \\
 & Wanda & -- & 39.50 & 22.92 & 61.15 & 39.94 & 19.88 & 52.17 & 14.00 & 35.01 \\
 & Wanda & ADAM & 205.82 & 25.63 & 57.02 & 34.13 & 17.66 & 50.51 & 13.00 & 32.99 \\
 & \cellcolor{shade}Wanda & \cellcolor{shade}\ours & \cellcolor{shade}35.44 & \cellcolor{shade}23.02 & \cellcolor{shade}61.64 & \cellcolor{shade}\textbf{42.93} & \cellcolor{shade}19.11 & \cellcolor{shade}50.12 & \cellcolor{shade}14.60 & \cellcolor{shade}35.24 \\
 & SparseGPT & SparseGPT & 36.88 & 23.00 & 61.97 & 40.99 & 19.71 & 53.59 & 14.60 & 35.64 \\
 & SparseGPT & ADAM & 224.34 & 23.15 & 56.75 & 35.65 & 17.49 & 47.36 & 12.20 & 32.10 \\
 & \cellcolor{shade}SparseGPT & \cellcolor{shade}\ours & \cellcolor{shade}35.61 & \cellcolor{shade}\textbf{23.85} & \cellcolor{shade}\textbf{62.35} & \cellcolor{shade}42.30 & \cellcolor{shade}\textbf{19.97} & \cellcolor{shade}\textbf{52.25} & \cellcolor{shade}\textbf{15.40} & \cellcolor{shade}\textbf{36.02} \\
\bottomrule
\end{tabular}
}
\caption{Comparison of \ours with other optimizers without convergence guarantees (ADAM). ADAM can lead to suboptimal solutions (Gemma 3 1B) or divergence of the model (OPT 125M).}
\label{tab:qp_vs_adam}
\end{table}

\autoref{tab:qp_vs_adam} illustrates this on Gemma 3 1B and OPT 125M \citep{opt} under 50\% unstructured sparsity. We show two examples in \autoref{tab:qp_vs_adam}, showing that ADAM results in suboptimal solutions. To further test the limitations of optimizers without convergence guarantees, we test ADAM on OPT-125M, and observe that it leads to divergence of the model. On Gemma 3 1B, ADAM yields competitive results in some cases (e.g., slightly lower perplexity for SparseGPT+ADAM at 27.12 versus \our's 27.35), but \ours achieves higher overall accuracy (e.g., 44.01\% for Wanda+\ours versus 43.72\% for Wanda+ADAM). However, on smaller models like OPT 125M, ADAM exhibits instability, leading to divergence and much higher perplexity (e.g., 205.82 for Wanda+ADAM versus 35.44 for Wanda+\our). This underscores the risks of using non-specialized optimizers for our column-wise QPs, where suboptimal or unstable solutions can degrade model quality. \our's use of a provably convergent solver with a per-column certificate, backed by the fallback, makes it the more robust choice for post-training pruning.

\subsection{Recovery fine-tuning on the calibration data}
\label{app:finetuning}

A natural alternative to a principled weight update is light recovery fine-tuning of the pruned model on the calibration data. We ran this with the same 128 C4 samples as fine-tuning data, ADAM with default HuggingFace Trainer settings, masks retained, and iteration counts swept over $\{1, 2, 4, 8, 16, 32, 64, 128\}$ on a single H100. \autoref{tab:adam_finetuning} reports perplexity and six-task average accuracy on LLaMA 3.2 1B and 3B at 50\% unstructured sparsity.

In the calibration-only regime of one-shot pruning, additional optimization on the same calibration data is not a competitive recovery mechanism. The sweep shows a clear transition from negligible gains to severe degradation as iterations grow. Accuracy peaks within 1 to 8 iterations (on 3B the single-iteration point is best for both selectors), where fine-tuning recovers at most $+0.17$ points over the corresponding pre-fine-tuning baseline on the 1B model and nothing on the 3B model (49.96\% vs.\ 49.95\% with Wanda; 50.20\% vs.\ 50.20\% with SparseGPT); it declines gently through 16 and 32 iterations, then collapses between 32 and 64 (perplexity 23.10 to 479.84 on 1B with Wanda masks). Across the sweep, calibration-set fine-tuning never approaches \ours before overfitting sets in: at its best iteration per setting, ADAM remains 0.20 to 1.41 points below \ours on the six-task average. Notably, SparseGPT + ADAM (8) on 3B attains the best sparse-model perplexity in the table (12.15, vs.\ 12.32 for SparseGPT and 12.43 for \ours) while trailing \ours by 1.22 accuracy points, a further instance of the perplexity/accuracy divergence discussed in \autoref{sec:results}.

Two caveats apply. First, this comparison is data-matched, not compute-matched: making fine-tuning benefit from \ours's compute budget would require substantially more data, which changes the problem setting. \ours targets the regime where only about 260K calibration tokens exist, while sparse retraining needs billions of tokens \citep{sparse_ft}. Second, memory: full-precision ADAM fine-tuning of LLaMA 3.1 8B does not fit on a single H100 even with gradient checkpointing (optimizer states plus gradients), and memory-reduced variants (8-bit optimizers, LoRA) alter the optimization problem and are out of scope here; \ours prunes the same model in 60 GB.

\begin{table}[tb]
\centering
\small
\begin{tabular}{l cc cc}
\toprule
\multirow{2}{*}{\textbf{Mask / Weight Update}} & \multicolumn{2}{c}{\textbf{LLaMA 3.2 1B}} & \multicolumn{2}{c}{\textbf{LLaMA 3.2 3B}} \\
\cmidrule(lr){2-3}\cmidrule(lr){4-5}
 & Perplexity & Avg. (\%) & Perplexity & Avg. (\%) \\
\midrule
Dense & 9.75 & 49.09 & 7.81 & 57.95 \\
\midrule
Wanda, no update & 23.51 & 40.01 & 12.92 & 49.95 \\
\cellcolor{shade}Wanda + \ours & \cellcolor{shade}18.84 & \cellcolor{shade}\textbf{41.33} & \cellcolor{shade}12.24 & \cellcolor{shade}\textbf{51.37} \\
Wanda + ADAM (1) & 23.50 & 39.95 & 12.92 & 49.96 \\
Wanda + ADAM (2) & 22.82 & 40.02 & 12.76 & 49.88 \\
Wanda + ADAM (4) & 22.70 & 40.06 & 12.68 & 49.92 \\
Wanda + ADAM (8) & 22.65 & 40.08 & 12.64 & 49.94 \\
Wanda + ADAM (16) & 22.78 & 40.04 & 12.71 & 49.90 \\
Wanda + ADAM (32) & 23.10 & 39.99 & 12.80 & 49.85 \\
Wanda + ADAM (64) & 479.84 & 37.68 & 80.39 & 46.08 \\
Wanda + ADAM (128) & 480.38 & 37.65 & 265.49 & 42.87 \\
\midrule
SparseGPT + own update & 18.84 & 42.35 & 12.32 & 50.20 \\
\cellcolor{shade}SparseGPT + \ours & \cellcolor{shade}18.09 & \cellcolor{shade}\textbf{42.72} & \cellcolor{shade}12.43 & \cellcolor{shade}\textbf{51.39} \\
SparseGPT + ADAM (1) & 18.83 & 42.32 & 12.32 & 50.20 \\
SparseGPT + ADAM (2) & 18.58 & 42.44 & 12.23 & 50.09 \\
SparseGPT + ADAM (4) & 18.50 & 42.49 & 12.18 & 50.14 \\
SparseGPT + ADAM (8) & 18.46 & 42.52 & 12.15 & 50.17 \\
SparseGPT + ADAM (16) & 18.55 & 42.46 & 12.20 & 50.11 \\
SparseGPT + ADAM (32) & 18.70 & 42.38 & 12.26 & 50.04 \\
SparseGPT + ADAM (64) & 112.27 & 39.35 & 51.49 & 47.40 \\
SparseGPT + ADAM (128) & 537.94 & 37.19 & 249.64 & 43.51 \\
\bottomrule
\end{tabular}
\caption{Recovery fine-tuning with ADAM on the 128 calibration samples, with the mask held fixed, for 1 to 128 iterations (in parentheses), at 50\% unstructured sparsity. Avg.\ is the six-task average accuracy.}
\label{tab:adam_finetuning}
\end{table}

\FloatBarrier
\section{Solver-tolerance ablation}
\label{app:tolerance}

\autoref{sec:results} uses a relative and absolute rAPDHG tolerance of $10^{-2}$. \autoref{tab:tolerance} compares it with tighter tolerances of $10^{-3}$ and $10^{-4}$ on LLaMA 3.2 1B and Gemma 3 1B at 50\% unstructured sparsity for all three mask selectors. Tightening the tolerance by one to two orders of magnitude changes the six-task average by at most about 0.6 points with no consistent direction, so the default sits at a good compute/quality trade-off, consistent with \autoref{sec:limitations}.

\begin{table}[tb]
\centering
\small
\begin{tabular}{l l ccc}
\toprule
\textbf{Model} & \textbf{Mask} & \textbf{$10^{-2}$ (default)} & \textbf{$10^{-3}$} & \textbf{$10^{-4}$} \\
\midrule
LLaMA 3.2 1B & Wanda & 18.84 / 41.33 & 18.63 / 41.56 & 18.64 / 41.27 \\
LLaMA 3.2 1B & SparseGPT & 18.09 / 42.72 & 17.91 / 42.21 & 17.86 / 42.12 \\
LLaMA 3.2 1B & Thanos & 18.77 / 41.94 & 18.79 / 42.30 & 18.38 / 42.54 \\
Gemma 3 1B & Wanda & 28.90 / 44.01 & 28.86 / 43.60 & 29.07 / 43.64 \\
Gemma 3 1B & SparseGPT & 27.35 / 43.76 & 27.05 / 43.47 & 27.08 / 43.51 \\
Gemma 3 1B & Thanos & 28.14 / 44.05 & 27.97 / 43.66 & 27.55 / 43.78 \\
\bottomrule
\end{tabular}
\caption{Solver-tolerance ablation at 50\% unstructured sparsity. Each cell reports WikiText2 perplexity / six-task average accuracy (\%) for the given relative and absolute rAPDHG tolerance.}
\label{tab:tolerance}
\end{table}

\FloatBarrier
\section{Scaling to LLaMA-3.1-70B and baseline pruning time}
\label{app:70b}

To show that the approach reaches deployment-relevant scales, we pruned LLaMA-3.1-70B with Wanda masks on a single H100. \autoref{tab:llama70b} reports the results. At 50\% unstructured sparsity, perplexity moves from 5.70 to 5.68 and the six-task average from 67.04\% to 67.30\%; at 60\%, perplexity moves from 8.04 to 8.26 while the average improves from 60.74\% to 61.37\% ($+0.63$ points, including MMLU 59.41\% to 60.73\% and WinoGrande 72.85\% to 74.03\%);
 the dense reference is 2.81 perplexity and 70.35\% average accuracy. The run took about 350 hours on a single H100, consistent with the growth of pruning time with model size reported in \autoref{sec:results}; we report only measured single-GPU numbers.

\autoref{tab:pruning_time} reports the mask-selection time of Wanda and SparseGPT under the identical protocol next to \ours's end-to-end time on a single H100. Mask selection is below 1\% of the total, so \ours's runtime is dominated by the QP solves.

\begin{table}[tb]
\centering
\small
\begin{tabular}{l l l cc}
\toprule
\textbf{Sparsity} & \textbf{Mask Selection} & \textbf{Weight Update} & \textbf{Perplexity} & \textbf{Avg. (\%)} \\
\midrule
Dense & -- & -- & 2.81 & 70.35 \\
\midrule
50\% unstructured & Wanda & -- & 5.70 & 67.04 \\
50\% unstructured & \cellcolor{shade}Wanda & \cellcolor{shade}\ours & \cellcolor{shade}5.68 & \cellcolor{shade}\textbf{67.30} \\
\midrule
60\% unstructured & Wanda & -- & 8.04 & 60.74 \\
60\% unstructured & \cellcolor{shade}Wanda & \cellcolor{shade}\ours & \cellcolor{shade}8.26 & \cellcolor{shade}\textbf{61.37} \\
\bottomrule
\end{tabular}
\caption{LLaMA-3.1-70B pruned with Wanda masks on a single NVIDIA H100 (about 350 hours end-to-end). Avg.\ is the six-task average accuracy.}
\label{tab:llama70b}
\end{table}

\begin{table}[tb]
\centering
\caption{Mask-selection time of Wanda and SparseGPT versus \ours's end-to-end pruning time on a single NVIDIA H100, under the identical protocol (128 calibration samples of 2048 tokens); mask selection is below 1\% of \ours's time.}
\label{tab:pruning_time}
\small
\begin{tabular}{lccc}
\toprule
\textbf{Model} & \textbf{Wanda (s)} & \textbf{SparseGPT (s)} & \textbf{\ours (h)} \\
\midrule
LLaMA 3.2 1B & 15 & 65 & 2.5 \\
Gemma 2 2B & 21 & 120 & 5.5 \\
LLaMA 3.2 3B & 27 & 155 & 7.0 \\
LLaMA 3.1 8B & 50 & 308 & 40.0 \\
LLaMA 3.1 70B & -- & -- & $\approx$350 \\
\bottomrule
\end{tabular}
\end{table}

\FloatBarrier
\section{Extended related work}
\label{app:related_work}

\autoref{sec:related_work} condenses this discussion.

Pruning compresses a trained network by removing weights that contribute little to its outputs, reducing both compute and memory at inference time. Work in this space has historically split along two lines: layer-wise methods, built on the Optimal Brain Surgeon (OBS) framework \citep{obs} applied per layer \citep{obc}, and global or end-to-end methods, exemplified by Optimal Brain Damage (OBD) \citep{obd}. We review both in turn, then cover orthogonal compression techniques.

\textbf{Layer-wise pruning.} Layer-wise methods prune one layer at a time, under the assumption that keeping each layer's output close to the dense model's will keep the end-to-end error small. OBS applied per layer \citep{obc} identifies the lowest-saliency weight in each layer and updates the remaining weights to compensate for its removal, but the full second-order machinery is too expensive to run on billion-parameter LLMs. Optimal BERT Surgeon \citep{obs_bert} shows that OBS-style approximations can be scaled to transformer-size models via block-diagonal Hessian approximations; \ours's shared-Hessian column-wise QP is a different decomposition of the same underlying layer-wise objective. SparseGPT \citep{sparsegpt} recasts the update as a sparse regression and solves it approximately, which scales to LLM sizes at the cost of optimality. Thanos \citep{thanos} follows the same recipe but updates several columns at a time and reports lower approximation error. DSnoT \citep{dsnot} adjusts the mask iteratively without weight training, which is an orthogonal design point to \ours's fixed-mask joint weight reconstruction. Exact solvers of the same post-mask convex problem exist: \citet{boza_admm} use ADMM, and ALPS \citep{alps} refines its ADMM-selected support with a vectorized GPU conjugate-gradient solve on the shared Hessian; both exploit the per-matrix shared Hessian. FISTAPruner \citep{fistapruner} relaxes the mask with an $\ell_1$ penalty solved by FISTA, and CHITA \citep{chita} jointly selects the support and updates the weights under a quadratic model. Wanda++ \citep{wandapp} composes a Wanda-style score with a block-level regional update. \appref{app:admm} compares \ours with ADMM head to head: the two reach essentially the same solution, ADMM is faster, and \ours differs in its certified per-column termination and its drop-in use on Wanda, SparseGPT, Thanos, ProxSparse, and ADMM masks.

\textbf{End-to-end pruning.} End-to-end methods use global information, typically derived from second-order gradients, to decide which weights to remove. OBD \citep{obd} is the classical example; it relies on a diagonal Hessian approximation, which the original, network-wide OBS \citep{obs} refined with the full inverse Hessian at a much higher cost. WoodFisher \citep{woodfisher} approximates the inverse Hessian with a Woodbury-based empirical Fisher, which brings the cost down but remains impractical at LLM scale \citep{mkor}; LLM Surgeon \citep{llmsurgeon} reaches LLM scale with Kronecker-factored curvature, and SparseLLM \citep{sparsellm} and Elsa \citep{elsa} prune globally through decomposed or ADMM subproblems, the latter arguing that layer-wise reconstruction itself becomes the bottleneck at high sparsity. MaskLLM \citep{maskllm} sidesteps second-order information altogether and casts mask selection as a classification problem trained with AdamW \citep{adamw}, reaching strong results at 2:4 sparsity. ProxSparse \citep{proxsparse} cuts MaskLLM's training cost by swapping the classifier for a proximal regularizer. Both still require optimization loops that are heavier than a single forward pass over calibration data.

\textbf{Other compression techniques.} Pruning is one tool among several and is often stacked with others. Quantization lowers the bit-width of weights and activations \citep{quantization_survey, quantization_survey2}. Low-rank adaptation \citep{slim, lqlora, slope} stores weights as products of smaller matrices. Knowledge distillation \citep{knowledge_distillation} trains a smaller student from a larger teacher. These techniques target different sources of redundancy than pruning and can be layered on top of it.

\FloatBarrier
\section{Language model usage in the paper}
\label{app:llm_usage}

Apart from the language models that are the subject of the experiments, language models were used only to improve the clarity of the writing, correct grammatical and typographical errors, and check adherence to the ACML author guidelines; they were not applied to any other component of this work.

\section{Reproducibility statement}
We have taken several measures to ensure the reproducibility of our results. The source code and scripts for reproducing all experiments are available at \url{https://github.com/paramathic/optima}. The main text (\autoref{sec:methodology} and \autoref{sec:results}) describes our method and experimental setup in detail, while \autoref{app:implementation} specifies implementation details, hyperparameters, and model configurations. Together, these resources ensure that independent researchers can reproduce our findings with minimal effort.

\fi

\end{document}